\documentclass[letterpaper]{article} 
\usepackage[]{aaai2026}  
\usepackage{times}  
\usepackage{helvet}  
\usepackage{courier}  
\usepackage[hyphens]{url}  
\usepackage{graphicx} 
\usepackage{natbib}  
\usepackage{caption} 
\usepackage[ruled,linesnumbered,vlined]{algorithm2e}
\usepackage{subcaption}
\usepackage{amsmath}
\usepackage{booktabs,makecell, multirow, tabularx}
\usepackage[table]{xcolor}
\usepackage{colortbl}

\title{Tackling Resource-Constrained and Data-Heterogeneity in Federated Learning with Double-Weight Sparse Pack}

\author {
    Qiantao Yang\textsuperscript{\rm 1}, 
    Liquan Chen\textsuperscript{\rm 1, \rm 2}\thanks{Corresponding author},  
    Mingfu Xue\textsuperscript{\rm 3}, 
    Songze Li\textsuperscript{\rm 1}
}
\affiliations {
    \textsuperscript{\rm 1} School of Cyber Science and Engineering, Southeast University, China\\
    \textsuperscript{\rm 2} Purple Mountain Laboratories, China\\
    \textsuperscript{\rm 3} School of Communication and Electronic Engineering, East China Normal University, China\\
    \{yangqiantao, lqchen, songzeli\}@seu.edu.cn, mfxue@cee.ecnu.edu.cn
}

\begin{document}

\maketitle

\begin{abstract}
Federated learning has drawn widespread interest from researchers, yet the data heterogeneity across edge clients remains a key challenge, often degrading model performance. Existing methods enhance model compatibility with data heterogeneity by splitting models and knowledge distillation. However, they neglect the insufficient communication bandwidth and computing power on the client, failing to strike an effective balance between addressing data heterogeneity and accommodating limited client resources. To tackle this limitation, we propose a personalized federated learning method based on cosine sparsification parameter packing and dual-weighted aggregation (FedCSPACK), which effectively leverages the limited client resources and reduces the impact of data heterogeneity on model performance. In FedCSPACK, the client packages model parameters and selects the most contributing parameter packages for sharing based on cosine similarity, effectively reducing bandwidth requirements. The client then generates a mask matrix anchored to the shared parameter package to improve the alignment and aggregation efficiency of sparse updates on the server. Furthermore, directional and distribution distance weights are embedded in the mask to implement a weighted-guided aggregation mechanism, enhancing the robustness and generalization performance of the global model. Extensive experiments across four datasets using ten state-of-the-art methods demonstrate that FedCSPACK effectively improves communication and computational efficiency while maintaining high model accuracy.
\end{abstract}

\begin{links}
    \link{Code}{https://github.com/NigeloYang/FedCSPACK}
\end{links}

\section{Introduction}

Federated learning (FL) has attracted attention for its ability to leverage the massive data of clients while protecting privacy by sharing model parameters between clients without exchanging raw data for collaborative training of models~\cite{McMahanMRA16, AlamLYZ22}. However, the data generated and collected by edge client devices across different geographical locations exhibit significant heterogeneity, resulting in Non-Independent and Identically Distributed (Non-IID) characteristics in their data. Its computing, storage, bandwidth, and other resources also vary, which causes heterogeneity in system resources. In this context, a single global model struggles to encompass all client-side local datasets, and traditional methods using a single global model often result in significant deviations in the distribution of local data across clients, leading to slow convergence, poor inference performance, and even the inability to deploy in resource-constrained environments~\cite{LiSAV2021, LiDCH22, OhKY22}. To solve the problem of heterogeneity in FL, some works propose personalized federated learning (PFL)~\cite{HuangCZWLPZ21, TanYCY22}, which tailors a local model for each client according to their data distribution and system resources to enhance the model's ability to solve the heterogeneity problem.

Existing works demonstrate that the essence of model performance degradation is client drift caused by data heterogeneity. When significant aggregation discrepancies exist between the local model and global model, the aggregated global model fails to adapt to the heterogeneous data distributions across clients adequately. To reduce client drift,~\cite{KarimireddyKMRS20} uses the difference between the old and global rounds of local models to compensate for the current round of local updates.~\cite{LiSZSTS20, AcarZNMWS21, ZhangLLX0DW22} add regularization to the local loss function to facilitate the local model to be close to the global model. There are also some model aggregation strategies~\cite{JunW22, ZhangHWSXMG23} and knowledge distillation~\cite{TanLLZ00Z22, XuTH23}. While these works improve the performance of models in the face of data heterogeneity, they overlook the problem of limited client resources.

In FL, clients are typically composed of physical devices with varying processors, memory, and bandwidth, which creates resource differences between devices (system resource heterogeneity). Since FL relies on frequent model interaction, limited resources clients often struggle to keep pace with the collaborative training of a complex global model. This leads to communication bottlenecks, computational delays, and participation imbalance~\cite{abs-1812-07210, Diao0T21}. Some works reduce resource constraints and improve the interaction efficiency and overall performance of the model at both ends by sharing sparse model parameters, optimizing model architecture, and model splitting~\cite{JiangXXWLQQ24, HorvathLALVL21, IsikPGWZ23, ZhangHCWSXMG23, WuLNZ023}. However, these works only consider the inconsistency of limited system resources, lacking consideration of the data heterogeneity of clients. In the real world, data heterogeneity and client resource constraints are not isolated but rather intertwined, synergistic core challenges. Data heterogeneity impacts model convergence speed and performance, while system resources constrain communication and computational efficiency. Achieving an effective balance between these two factors can reduce the burden on limited-resource clients while improving model performance, thereby promoting the viable deployment of FL in the real world.

Given the limitations of existing works, we proposed a cosine sparsification parameter packing and dual-weighted aggregation method for PFL, which effectively balances data heterogeneity and limited system resources, aiming to reduce communication bottlenecks while maintaining model performance. Specifically, the client divides the model parameters into multiple flat parameter packages and adaptively selects the K parameter packages with the highest similarity through cosine similarity for sharing. This reduces communication overhead during transmission and effectively helps resource-constrained clients overcome communication bottlenecks. When the client computes the shared parameter packet, it generates a mask matrix that anchors the parameter packet and assigns a dual weight value to the mask, consisting of a direction weight and a distribution distance weight, helping the server to effectively align the parameter packet. During the aggregation phase, the server uses the dual weight values of the mask to complete the weighted aggregation, which accelerates the global model absorption of new knowledge and mitigates the negative impact of client data heterogeneity on the training process. To verify the effectiveness of FedCSPACK, we conduct extensive experiments across various heterogeneous scenarios. The results demonstrate that FedCSPACK can effectively balance data heterogeneity and resource constraints while ensuring model performance and client communication efficiency. The main contributions of our work are as follows:

\begin{itemize}
\item To the best of our knowledge, FedCSPACK is the first personalized federated learning method implemented at the package level, which achieves a balance between limited client resources and data heterogeneity through parameter packaging and dual-weight aggregation.

\item Packaging local model parameters effectively reduces communication overhead. A personalized mask is generated based on the parameter package, and a dual weight is calculated for the parameter package using KL divergence and cosine similarity, which improves the server's aggregation efficiency and knowledge absorption capacity for scattered parameter packages, mitigating the impact of data heterogeneity on model performance.

\item We compare FedCSPACK with 10 state-of-the-art (SOTA) methods and conduct extensive evaluations on real-world datasets with different data heterogeneity. The results show that FedCSPACK improves training speed by 2-5$\times$ and model accuracy by 3.34\%, while maintaining computational efficiency.
\end{itemize}

\section{Relate Works}
FL may not be as good as the individual training performed locally by each client in a variety of heterogeneous scenarios~\cite{TanLML0022, LeeJSBY22}. To address these challenges, personalized federated learning was developed~\cite{HuangCZWLPZ21, YiWLSY23, ZhangHCWSXMG23}, which enables each client to train a PFL model that suits the clients.

In PFL, FedProx~\cite{LiSZSTS20} constrains the offset between the local model and the global model by introducing L2 into the local objective function, but the size of the L2 will affect the balance between the degree of the model and the convergence speed. FedNova~\cite{WangLLJP20} normalizes the global model based on the number of local steps for each party. FedGH \cite{YiWLSY23} uploads the calculated local average representation and category labels to the server, which trains the global prediction header and broadcasts it to the client, which replaces the local prediction header. FedDBE~\cite{ZhangHCWSXMG23} reduces the domain differences between the two in the representation space by facilitating bidirectional knowledge transfer between the server and the client. FedAS \cite{YangHY24} enhances the localization of global parameters by combining them with local insights. However, these schemes do not deeply consider the division between model parameters, which can easily lead to bottlenecks in model performance. FedNTD~\cite{LeeJSBY22} uses global models and historical models as teachers to promote comprehensive knowledge distillation and complete the transfer of global generalized knowledge and historical personalized knowledge to local models, thereby mitigating catastrophic forgetting. FedPAC \cite{XuTH23} improves the accuracy and robustness of the model as a whole through feature alignment and classifier collaboration, it can be challenging to implement and optimize the model for resource-constrained clients. Li et al.~\cite{LiHS21} corrected the user model optimization direction by comparing the losses of the global model in two rounds of training. FedALA~\cite{ZhangHWSXMG23} fuses the global model of each round with the local model of the previous round to preserve the local personalized information.

Additionally, there are the Top-k-based model-sparse methods, e.g., FedSPU~\cite{NiuDQ25}. Although these methods effectively solve heterogeneous problems or resource constraints in FL, they only consider one of two factors separately, ignoring the challenges of coordination between data heterogeneity and limited system resources.
\begin{figure*}[ht]
  \centering
  \includegraphics[height=0.4\linewidth,width=.95\textwidth]{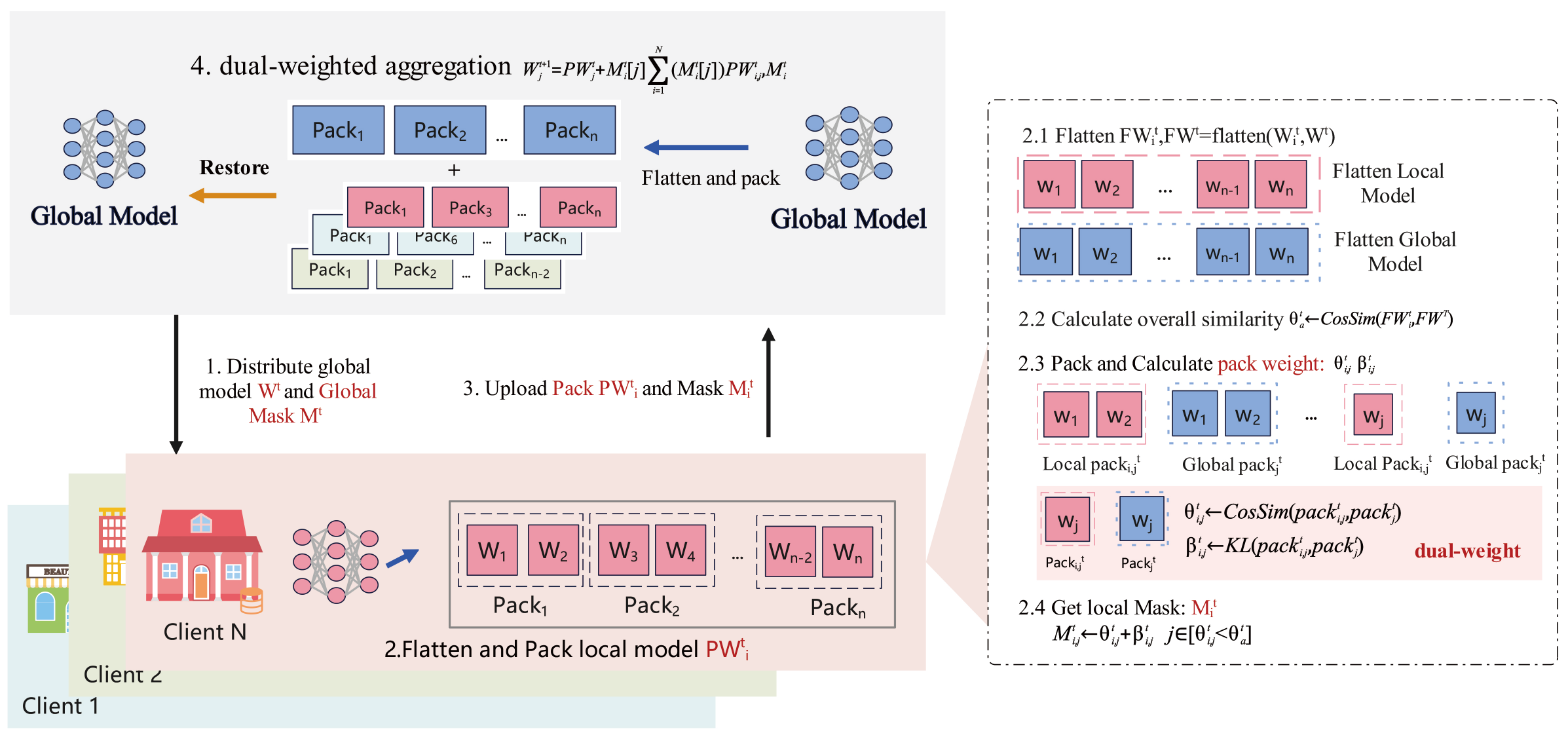}
  \caption{\textbf{FedCSPACK Overview.}~The client shows the detailed process of the generation and selection of the local parameters package and mask. The server side displays the details of the dual-weighted aggregation.}
  \label{figure:fig1}
\end{figure*}
FedCSPACK aims to improve the computational efficiency of FL on resource-constrained devices and alleviate communication bottlenecks. Through a dual-weight aggregation mechanism, FedCSPACK reduces the impact of data heterogeneity on client and server model performance, achieving efficient and robust personalized model training.

\section{Problem Formulation}
In FL, suppose $N$ clients are participating in training, their local dataset is ${D_1, D_2, ..., D_N}$ and the local model is ${w_1, w_2, ..., w_N}$, and the training goal of the collaboration is to minimize the average loss of the global model $F(w)$ on all client data:

\begin{align} \label{pack-eq1} 
min F(w) = \sum_{i=1}^{N}\frac{\left|D_i\right|}{\sum \left|D_i\right|} L_i(F(w),D_i)
\end{align}

\noindent where the parameter $w$ of the global model is a $d$-dimensional vector. $L_i(F(w),D_k)$ is the loss function of the global model $F(w)$ on the local dataset $D_i$ on client $i$. $\left| D_i\right|$ is the dataset size for client $i$, $sum \left| D_i \right|$ is the sum of all client dataset sizes.

Because client data is the Non-IID distribution, the best global model obtained through training does not guarantee the best generalization performance across all clients. The goal of the PFL is to determine the optimal set of local models ${w_1^ast, w_2^ast, ..., w_N^ast}$, allowing multiple models to coexist, ensuring that each client $i$ learns an optimal personalized model based on its local goals:

\begin{align} \label{pack-eq2} 
w_1^\ast,...,w_N^\ast = \underset{w_1,...,w_N}{argmin} \sum_{i=1}^{N}\frac{\left|D_i\right|}{\sum \left|D_i\right|} L_i(F_i(w_i),D_i)
\end{align}

\noindent where $w_i^{\ast}$ is the optimal model for client $(i\in N)$, and $L_i(F_i(w_i), D_i)$ is the loss function for client $i$. Considering that data heterogeneity and limited system resources coexist in the real scenario, FedCSPACK can effectively obtain the best local model and minimize the overall experience loss through parameter packaging and double-weighted aggregation, thereby achieving a balance between limited client resources and data heterogeneity.

\section{FedCSPACK Design}

\subsection{Overview}
To address the above challenges, we propose FedCSPACK, which efficiently utilizes limited resources by packaging model parameters and selectively sharing partial parameter packages, while enhancing the model’s compatibility with heterogeneous data via a double-weight mask aggregation mechanism. The framework of the FedCSPACK is illustrated in fig.~\ref{figure:fig1}. In the $t$-round communication, FedCSPACK's steps are as follows:

\begin{itemize}
\item \textbf{Step 1.} The server completes the aggregation process of the global model $W^t$ and the global mask $M^t$, and then broadcasts them to all clients.

\item \textbf{Step 2-3.} Client $i$ updates the local model $W_i^t$ on the local data $D_i$ under the guidance of $M^t$ and $W^t$. Before sharing $W_i^t$, client $i$ will convert $W_i^t, W^t$ into one-dimensional vectors $FW_i^t, FW^t$ and use cosine similarity to calculate the overall similarity $\theta_a^t$ between the two models. Then, according to the size of $PACK$, $FW_i^t$ and $FW^t$ are packaged into parameter packages $PW_{i,j}^t$ and $PW_j^t$, and the similarity threshold $\theta_{i,j}^t$ and the discrete value $\beta_{i,j}^t$ of each $PW_{i,j}^t$ and $PW_j^t$ are calculated. Meanwhile, use Top-k to select K $PW_{i,k}^t,k<j$ by $\theta_{i,j}^t < \theta_a^t$ as the client $i$ shared parameter pack $PW_{i,k}^t$. Finally, to effectively align the $PW_{i,k}^t$ of client $i$, let the valid weight value of the mask $M_{i,j}^t = \theta_{i,j}^t + \beta_{i,j}^t$.

\item \textbf{Step 4.} The server collects local model $W_i^t$ and local mask $M_i^t$ shared by client $i$, integrates $M_i^t$, and completes the alignment-weighted aggregation task of the new round of global model $W^{t+1}$ according to its effective weight value.
\end{itemize}

Perform steps 1-4 above until the specified epoch or model $W_i$ converges for client $i\in [N]$. Details of the FedCSPACK implementation can be found in algorithm~\ref{algorithm：alg1}.

\begin{algorithm}
\KwIn{Local datasets $D_i$, Number of client N, Global epoch T, Local epoch E, Learning rate $\eta$}
\KwOut{Global Model $W^{T}$}

Server Executes: \\
Initialize Global Model $w^{t}$ \\
\For{global epoch $t \in [T]$}{
	Random Sample Clients Subset $S^t$ \\
	\For{each client $i \in S_t$ parallel}{
	   	$PW_{i,k}^t,M_i^t \gets$ \textbf{Client($i,W^t,M^t$)} \\
	}
	Update $W^{t+1},M^{t+1}\gets$ Eq.~\ref{align:a9}\\
	Send $W^{t+1},M^{t+1}$ to all clients \\
}
Return $W^T$ \\

Local Executes: \textbf{Client($i,W^t,M^t$)} \\
Receive Global Model: $W^t$ \\
Update: $W_i^t\gets update(W^t,M^t)$ \\

\For{local epoch $le \in [E]$}{
	$W_i^{t} \gets W_{i}^{t} - \eta\nabla f_i(W_{i}^{t})$ \\
}

$\Delta W_{i}^{t} \gets W^{t} - W_{i}^{t}$ \\

Flattened: $FW_i^t,FW^t \gets flattened(W_i^t,W^t)$\\
Get Mask and Parameters pack: $M_i^t,PW_{i,k}^t \gets $ Eqs.~\ref{align:a5} and \ref{align:a8} \\

Return $PW_{i,k}^t,M_i$ to the server \\

\caption{FedCSPACK}
\label{algorithm：alg1}
\end{algorithm}

\subsection{Top-k Based Cosine Parameter Packing}
Frequent model interactions between clients and the server pose a significant challenge to limited resource clients, as the frequent interaction of massive model parameters reduces their interaction efficiency and imposes a heavy communication burden on limited resource clients. Top-k \cite{AjiH17} is a commonly used communication optimization method, model parameters change dynamically during iterative training, and traditional Top-K sparsity methods find it difficult to select an appropriate parameter subset in each iteration. To alleviate the communication bottleneck and accurately select parameter subsets, we propose a packet-level Top-K parameters packaging and sharing mechanism guided by cosine similarity, because lower cosine similarities mean higher inference loss, and improving these similarities may lead to better model performance. 

After client $i$ receives global model $W^t$ and global mask $M^t$, it flattens the local model $W_i^t$ and $W^t$ to obtain $FW_i^t,FW^t$, then packages $FW_i^t,FW^t$ into $PW_i^t,PW^t$ according to $PACK$, and completes the local model update $PW_i^t[j]\gets PW^t[j],if M^t[j]=Valid$. Then, $PW_i^t[j]$ is restored and starts $t$-round local training. Before sharing a local model, client $i$ flattens $W_i^t$ and $W^t$ to obtain $FW_i^t,FW^t$, using cosine similarity (eq.~\ref{align:a3}) to calculate the overall similarity threshold $\theta_a^t$ between $FW_i^t$ and $FW^t$.

\begin{align} \label{align:a3} 
\theta_a^t = CosSim(FW_i^t,FW^t) = \frac{FW_i^t \cdot FW^t}{\parallel FW_i^t \parallel \parallel FW^t \parallel} 
\end{align}

Then, client $i$ will package $FW_i^t$ and $FW^t$ according to the size of $PACK$, ensuring that each package $PW_{i,j}^t \gets \frac{FW_i^t}{PACK}, j\in\left | \frac{FW_i^t}{PACK}\right |$ contains $PACK$ parameters (global model $PW_j^t \gets \frac{FW^t}{PACK}$), calculating the package similarity threshold $\theta_{i,j}^t$ according to eq.~\ref{align:a4}. 

\begin{align} \label{align:a4} 
\theta_{i,j}^t = \frac{PW_{i,j}^t \cdot PW_j^t}{\parallel PW_{i,j}^t \parallel \parallel PW_j^t \parallel},\; j \in \left|PW_i^t \right|
\end{align}

Next, client $i$ uses eq.\ref{align:a5} to select K packages ($PW_{i,k}^t,k\in j$) that is $\theta_{i,j}^t<\theta_a^t$ as the shared parameter packages of client $i$, and the rest of the parameter packets $PW_{i,j\setminus k}^t$ will become unique features of the client, which avoids the interference of the heterogeneity of the rest of the clients and maximizes the utility of each communication round. 

\begin{align} \label{align:a5} 
PW_{i,k}^t = TopK(PW_{i,j}^t,\theta_{i,j}^t,\theta_a^t),\; j \in\left|PW_i^t\right|
\end{align}

Although $PW_{i,k}^t$ improves communication efficiency, misaligned or erroneous aggregation of $PW_{i,k}^t$ can lead to biased global model updates, degrading overall model performance. To ensure that server efficiently indexes the position of $PW_{i,k}^t$ and effectively completes global model aggregation, we design a client mask $M_{i,j}^t$ to locate the position of $PW_{i,j}^t$ and considers $\theta_{i,k}^t \gets \theta_{i,j}^t < \theta_a^t$ as the valid index position of $PW_{i,k}^t$ and the proportion of the aggregate weight occupied according to eq.\ref{align:a6}.

\begin{align} \label{align:a6} 
M_{i,j}^t &= \begin{cases}
  \theta_{i,k}^t & \theta_{i,k}^t \in\theta_{i,j}^t \\
  0 & \theta_{i,j} \setminus \theta_{i,k}^t
\end{cases}
\end{align}

Through this packet-level model parameter selection and sharing mechanism, our method ensures that personalized models maintain model performance while communicating efficiently on resource-constrained clients.

\subsection{Mask Double Weight Aggregation}
Due to data heterogeneity, the global model aggregated by servers often struggles to adapt to the varying data distribution across clients. Although cosine similarity can maintain consistency between local updates and global directions in parameter space, it only reflects directional alignment. It cannot measure distance differences and magnitude shifts between parameter values. This insensitivity to magnitude can lead to overweighting the corresponding mask $M_{i,j}^t$, excessively influencing the aggregation results, and compromising model stability. To address this, we propose a dual-weight parameter pack aggregation mechanism. Based on directional weights based on cosine similarity, we further introduce the Kullback-Leibler (KL) divergence to quantify the distance shift between local and global model parameter updates. Specifically, When calculating the packet similarity threshold $\theta_{i,j}^t$ of $PW_{i,j}^t$ and $PW_{j}^t$, client $i$ will use KL divergence (eq.~\ref{align:a7}) to calculate the distance difference threshold $\beta_{i,j}^t$ between $PW_{i,j}^t$ and $PW_{j}^t$, and add $\beta_j$ as a supplement weight of $M_{i,j}^t$.

\begin{align} \label{align:a7} 
\beta_{i,j}^t = \sum PW_{i,j}^t \log\frac{PW_{i,j}^t}{PW_j^t},\; j\in\left|PW_i^t\right|
\end{align}

Then, we perform a new allocation process for the mask weights $M_{i,j}^t$ in eq.~\ref{align:a8} according to eq.~\ref{align:a6} to ensure that customers extract the weight ratio of each model parameter package more comprehensively. This strategy preserves the benefits of directional alignment while effectively mitigating the impact of distributional discrepancies, improving the common-sense ability of the global model to absorb each parameter package.

\begin{align} \label{align:a8} 
M_{i,j}^t &= \begin{cases}
  \theta_{i,k}^t + \beta_{i,k}^t & \theta_{i,k}^t \in\theta_{i,j} \\
  0 & \theta_{i,j} \setminus \theta_{i,k}^t
\end{cases}
\end{align}

After the server receives shared parameter packages $PW_{i,k}^t,i\in S^t$ and mask $ M_i^t,i\in S^t$, it first updates the global mask $M^{t+1}= sum(M_i^t)$ to obtain the total weight of each parameter packet position. Then, flatten the global model $W^t$ and complete the parameter packing ($PW_{i,k}^t$) according to the PACK size. Under the guidance of eq.~\ref{align:a9}, the server performs weighted aggregation on the client's parameter pack and fuses the new result $PW_{i,j}^t$ with $PW^t$ to obtain a new global model $PW_j^{t+1} \gets PW_j^t + PW_{i,j}^t$.
\begin{align} \label{align:a9} 
PW_{i,j}^t &=\begin{cases}
  \sum_{i=1}^{S_t} \frac{M_{i,j}^t} {M_j^{t+1}} PW_{i,k}^t &M_{i,j}^t \ne 0 \\
  0 & else
\end{cases}
\end{align}

As training progresses, the global model can efficiently and quickly absorb client model knowledge through mask double-weighted aggregation at the packet level, preserving its generalization capability. When clients merge a new global model, the stability and generalization performance of the global model under cross-client heterogeneous data are enhanced because the local parameter package retained by the client contains the local personalized knowledge.

\section{Experiments}

\subsection{Experiment Setting}
\noindent\textbf{Dataset Setup}: We use four publicly available image classification datasets as our experimental data: Fashion-MNIST (FMNIST), EMNIST, CIFAR-10, and CIFAR-100. To verify the effectiveness of \textbf{FedCSPACK}, we use CNN and ResNet-18 as the test model. To simulate this heterogeneity, we employ Dirichlet (Dir($\alpha$)) and pathological sampling, which are widely used in FL~\cite{SebastianPTJHVA18, AcarZNMWS21, LiDCH22}. A smaller $\alpha$ value corresponds to greater Non-IID characteristics. Dirichlet sampling is shown in fig.~\ref{figure:fig2}. Pathological Sampling and experimental setup information are in Appendix A. 
\begin{figure}[htbp]
  \centering
  \includegraphics[height=0.38\linewidth, width=.85\linewidth]{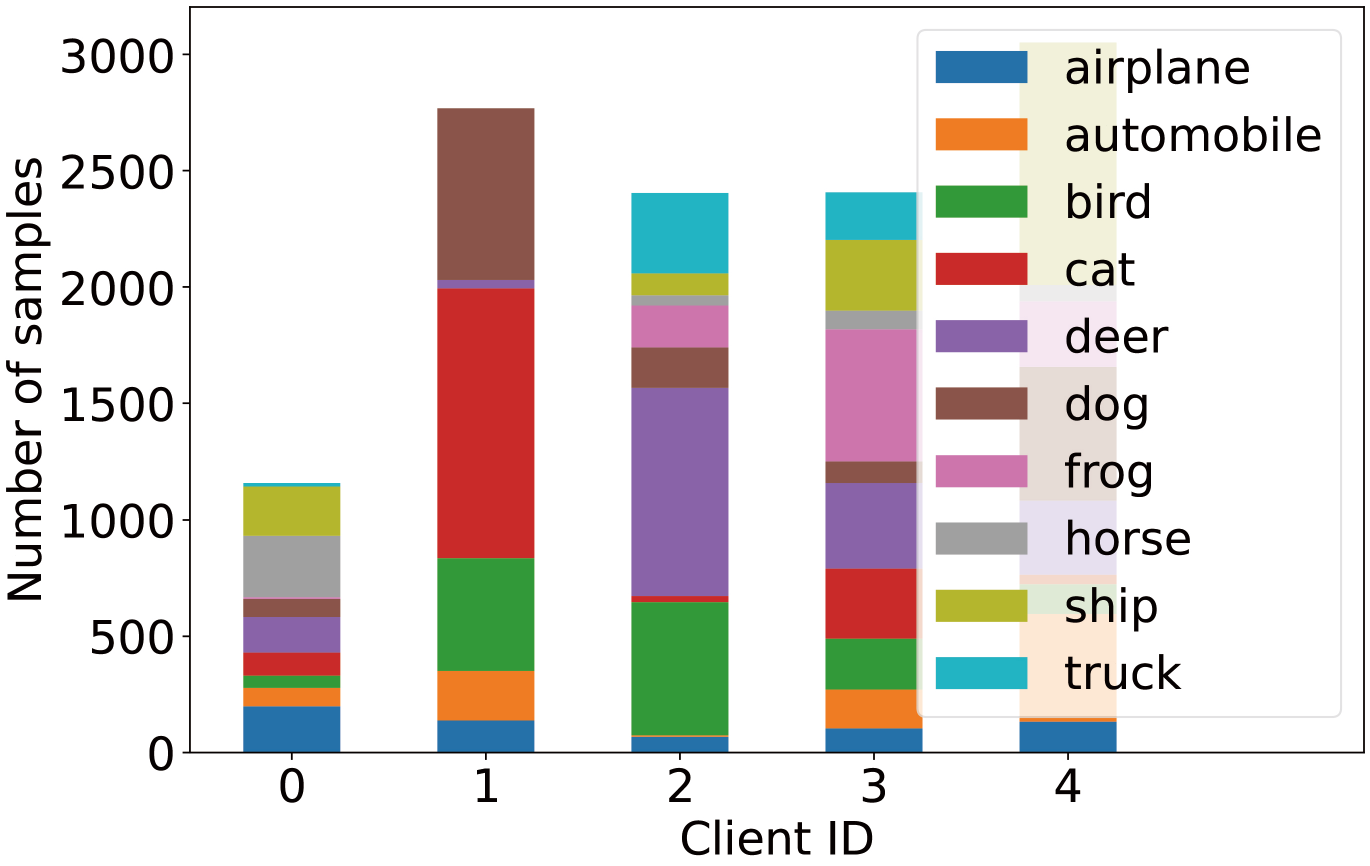}
  \caption{Data heterogeneity of 5 clients is simulated using dirichlet sampling ($Dir(0.6)$) on CIFAR-10.}
  \label{figure:fig2}
\end{figure}
All experiments are repeated three times, and the average value is taken as the experimental result.

\noindent\textbf{Baseline}: We compare FedCSPACK with ten SOTA methods. FedAvg~\cite{McMahanMRA16}, FedProx~\cite{LiSZSTS20}, and FedNova~\cite{WangLLJP20} are traditional methods; FedALA~\cite{ZhangHWSXMG23} in the aggregation scheme helps weak clients to choose the appropriate ratio of integration between the global and local models. Model Spliting (MOON \cite{LiHS21}, FedDBE~\cite{ZhangHCWSXMG23}, and FedAS~\cite{YangHY24}) and Knowledge Distillation (FedNTD~\cite{LeeJSBY22} and FedPAC~\cite{XuTH23}) can effectively improve accuracy and communication efficiency. FedSPU~\cite{NiuDQ25} improves communication efficiency by sparsifying model parameters.
\begin{table*}[htbp]
\centering
\small
\begin{tabular}{l| ccc | ccc | ccc | ccc}
\toprule
\multicolumn{1}{l}{DataSet} & \multicolumn{3}{c|}{FMNIST} & \multicolumn{3}{c|}{EMNIST} & \multicolumn{3}{c|}{CIFAR-10} & \multicolumn{3}{c}{CIFAR-100}\\ 
 \cmidrule{2-4}\cmidrule{5-7} \cmidrule{8-10} \cmidrule{11-13}
\multicolumn{1}{l}{Dirichlet($\alpha$)} & 0.3 & 0.6 & 1.0 & 0.3 & 0.6 & 1.0 & 0.3 & 0.6 & 1.0 & 0.3 & 0.6 & 1.0\\
\cmidrule{1-13}
FedAvg  & 84.39     & 86.96     & 87.65     & 83.12     & 84.03     & 83.98     & 69.71     & 74.71    & 75.01     & 39.15     & 40.44     & 40.34     \\
FedProx & 84.39     & 86.96     & 87.65     & 84.55     & 84.12     & 83.99     & 69.58     &{\bf74.59}& 74.54     & 38.48     & 39.58     & 39.68     \\
FedNova & 84.40     &{\bf 87.50}&{\bf 88.68}&{\bf 84.97}& 84.18     & 84.03     &{\bf 70.93}& 74.37    & 75.53     & 38.65     & 40.30     & 39.96     \\
MOON    & 85.44     & 86.40     & 87.90     & 83.17     & 84.08     & 83.94     & 70.03     & 74.63    &{\bf 76.00}& 38.43     & 39.85     & 39.90     \\
FedDBE  & 85.06     & 86.28     & 88.24     & 83.14     & 84.08     & 83.83     & 69.66     & 74.76    & 74.80     & 38.33     & 40.03     & 39.92     \\
FedAS   & 42.47     & 57.27     & 71.96     & 70.46     & 77.32     & 77.63     & 43.97     & 70.00    & 69.46     & 20.51     & 24.12     & 31.42     \\
FedNTD  & 84.35     & 86.67     & 88.01     & 83.56     &{\bf 84.49}&{\bf 84.32}& 70.32     & 74.68    & 75.52     &{\bf 39.44}&{\bf 40.67}&{\bf 40.88}\\
FedPAC  & 41.53     & 65.73     & 69.34     & 58.11     & 59.02     & 66.98     & 57.23     & 68.78    & 58.60     & 35.39     & 38.63     & 39.54     \\
FedALA  & 84.39     & 86.96     & 87.65     & 83.29     & 83.98     & 83.82     & 69.49     & 74.64    & 74.96     & 38.93     & 40.03     & 40.24     \\
FedSPU  &{\bf 85.29}& 88.76     & 88.25     & 83.56     & 84.22     & 83.91     & 67.38     & 74.52    & 74.62     & 37.81     & 38.25     & 39.35     \\
\cmidrule{1-13}
{\bf FedCSPACK(ours)} & {\bf 88.13} & {\bf 89.50} &{\bf 90.73} & {\bf 85.55} & {\bf 86.26} & {\bf 86.19} & {\bf 73.23} & {\bf 77.15} & {\bf 78.71} & {\bf 41.60} & {\bf 42.96} & {\bf 43.20}\\
\bottomrule
\end{tabular}
\caption{The test Top-1 accuracy (\%) across three Dirichlet sampling with $\alpha \in [0.3, 0.6, 1.0]$ on four datasets. The best result is FedCSPACK, while the bold values denote the best methods in SOTA.}
\label{Table:t1}
\end{table*}
\begin{table*}[ht]
\centering
\begin{tabular}{lc ccccccc}
\toprule
DataSet / Model                     & Metric  & FedAvg & Tradition & Split & Distillation & FedALA & FedSPU & {\bf FedCSPACK} \\
\hline
\multirow{2}{*}{FMNIST/CNN}         & Traffic & 0.30   & 0.30   & 0.30   & 0.32   & 0.30   & 0.07  & {\bf 0.02 } \\ 
				                    & Time    & 1.13   & 1.45   & 2.34   & 1.80   & 1.18   & 1.41  & {\bf 1.21 } \\
\hline				  
\multirow{2}{*}{EMNIST/CNN}         & Traffic & 18.18  & 18.18  & 18.18  & 18.29  & 18.18  & 4.29  & {\bf 0.73 } \\
				                    & Time    & 11.16  & 14.37  & 23.11  & 17.92  & 11.67  & 13.92 & {\bf 11.32 } \\
\hline
\multirow{2}{*}{CIFAR-10/CNN}       & Traffic & 2.59   & 2.59   & 2.59   & 2.61   & 2.59   & 0.61  & {\bf 0.20 } \\
                                    & Time    & 1.40   & 1.83   & 2.89   & 2.23   & 1.46   & 1.75  & {\bf 1.42 } \\
\hline				  
\multirow{2}{*}{CIFAR-100/ResNet18} & Traffic & 251.00 & 251.00 & 251.00 & 251.80 & 251.00 & 49.22 & {\bf 9.24 } \\
				                    & Time    & 0.81   & 1.05   & 1.67   & 1.30   & 0.84   & 1.01  & {\bf 0.88 } \\						
\bottomrule
\end{tabular}
\caption{On four datasets, with training rounds T=100 and Dir(0.3), the total training time (hours) and the number of parameters transmitted (GB) of FedCSPACK and SOTA are compared. Tradition: FedProx and FedNova. Split: MOON, FedDBE, and FedAS. Distillation: FedNTD and FedPAC. Parameter packing improves the computational and communication performance. }
\label{table:t2}
\end{table*}

\subsection{Performance Comparison}
\textbf{The Overall Performance of the Global Model}. Table~\ref{Table:t1} lists the results of FedCSPACK and other FL SOTA methods on the original dataset for basic models from shallow to deep. On FMNIST, as the data distribution shifts from $Dir(0.3)$ to $Dir(1.0)$, the model's accuracy generally improves. FedCSPACK performs well under all data distributions, especially at $Dir(1.0)$. It achieved an accuracy rate of 90.73\%, significantly higher than that of other methods. On complex CIFAR-10 and CIFAR-100, FedCSPACK reached 78.71\% and 43.20\% on $Dir(1.0)$, respectively. On EMNIST, the model's accuracy is generally high. This might be because the increase in data volume helps the model learn features better. As the data distribution shifted from $Dir(1.0)$ to $Dir(0.3)$, the model's accuracy declined. This might be due to the homogenization of data distribution leading to a reduction in samples of certain categories, thereby affecting the generalization ability of other scheme models. However, FedCSPACK can still ensure the high accuracy of the model through the dual-weight package aggregation method, especially reaching 86.26\% at $Dir(0.6)$ on EMNIST. The pathological result is in Appendix B. Overall, FedCSPACK can maintain a high model accuracy rate through dual-weight aggregation, effectively adapting to datasets ranging from simple to complex and from small to massive.
\begin{figure}[ht]
\centering
\includegraphics[height=0.6\linewidth, width=0.9\linewidth]{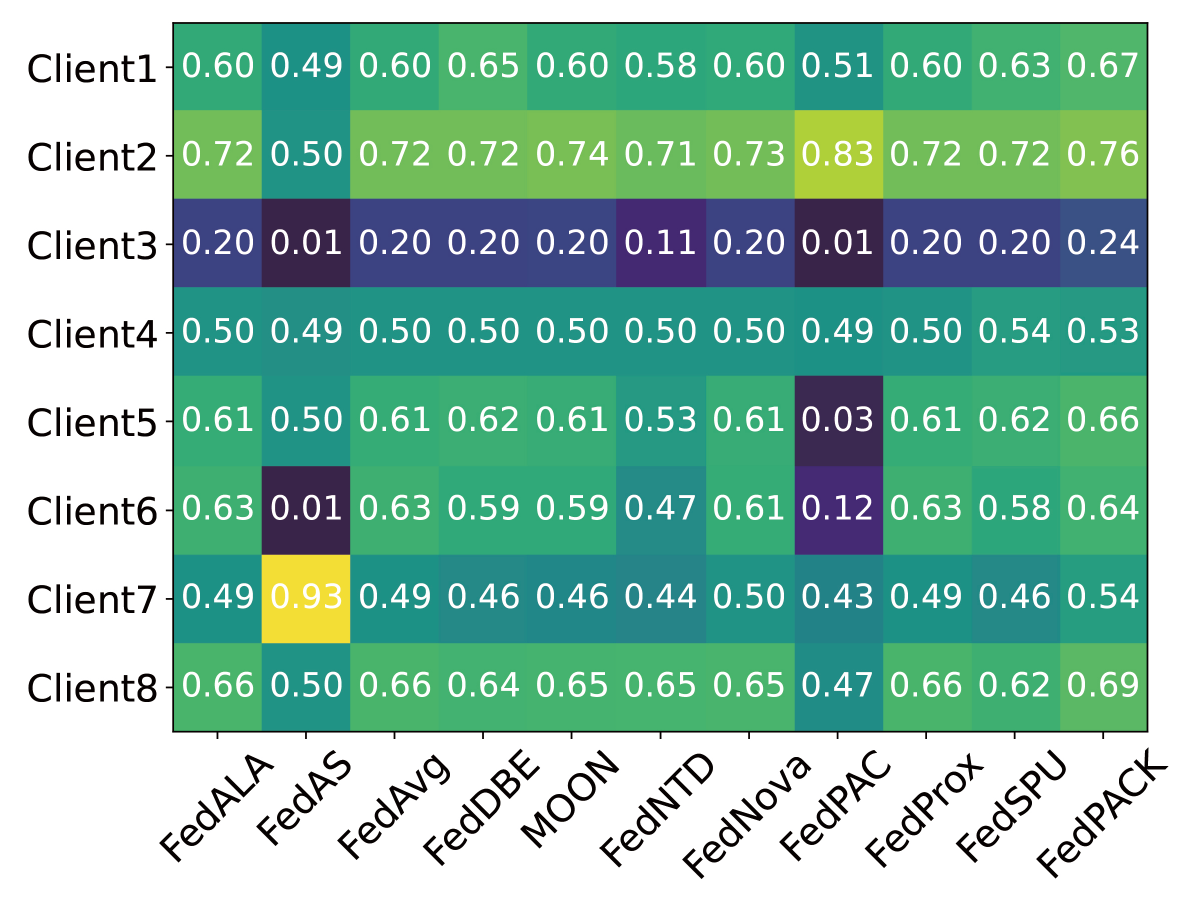}
\caption{The generalization of the global model on the clients in CIFAR-10 and Dir(0.3).}
\label{figure:fig3}
\end{figure}

\textbf{The Generalization Ability of the Global Model on the Client}. 
As shown in fig.~\ref{figure:fig3}, the global model trained by FedCSPACK shows effective model performance in adapting to the heterogeneous data of the client. On Client 5 with the best generalization performance, FedCSPACK achieves a model accuracy of 0.66, while the highest accuracy of other methods was 0.62, representing a 6.5\% improvement. On client 3 with the worst generalization performance, FedCSPACK's generalization performance remains stable, 20\% higher than the best SOTA (FedDBE and FedSPU). Although FedCSPACK's generalization performance is lower than FedPAC and FedAS on Client 2 and Client 7, it is not as good as FedCSPACK on other clients. This shows that FedCSPACK's use of Top-k cosine parameter packs and dual-weighted aggregation scheme can effectively ensure the generalization performance of the global model and make it more stable. More results are in Appendix C. 
\begin{figure*}[htbp]
  \centering
  \subfloat[CPR=1.0,Dir(0.5)]{
    \includegraphics[height=0.18\linewidth,width=.235\linewidth]{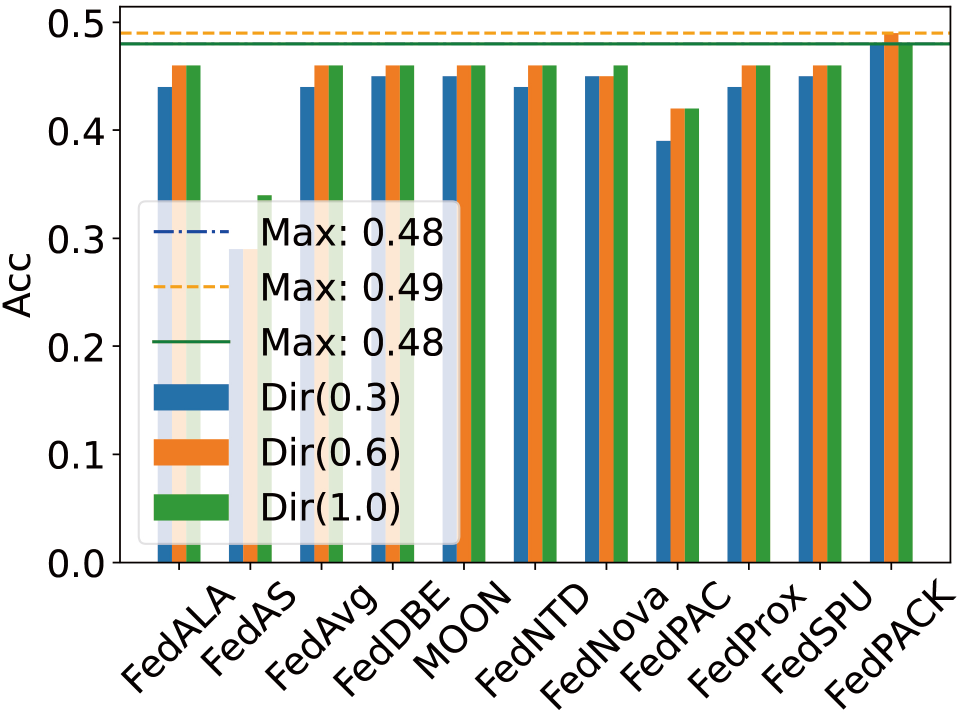}
    \label{fig4-cpr1}
  }
  \subfloat[CPR=0.6,Dir(0.5)]{
    \includegraphics[height=0.18\linewidth,width=.235\linewidth]{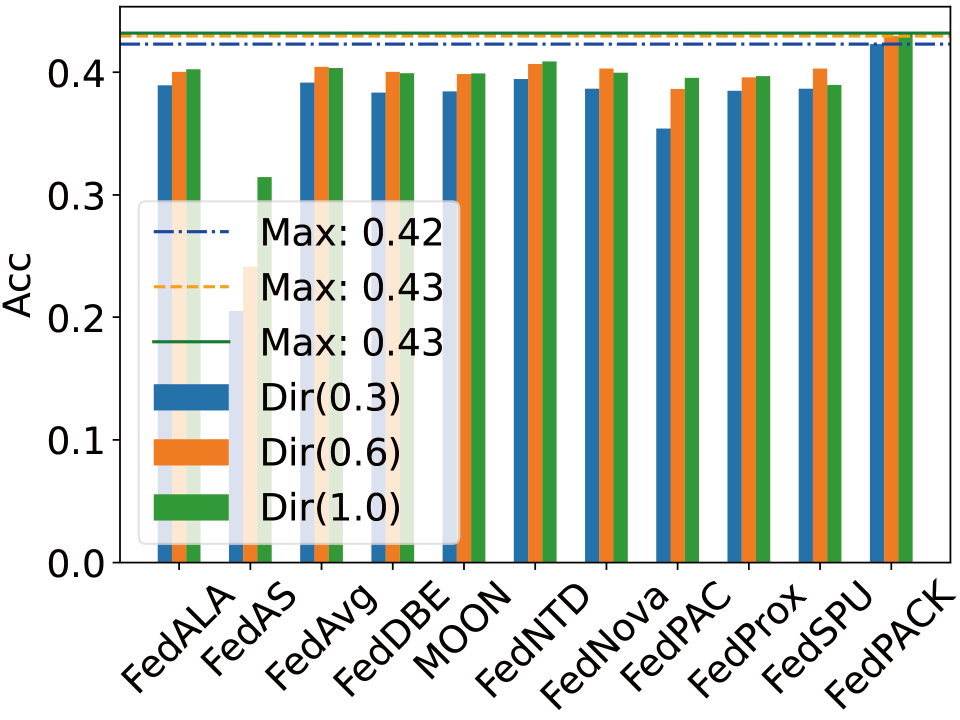}
    \label{fig4-cpr2}
  }
  \subfloat[CPR=0.3,Dir(0.5)]{
    \includegraphics[height=0.18\linewidth,width=.235\linewidth]{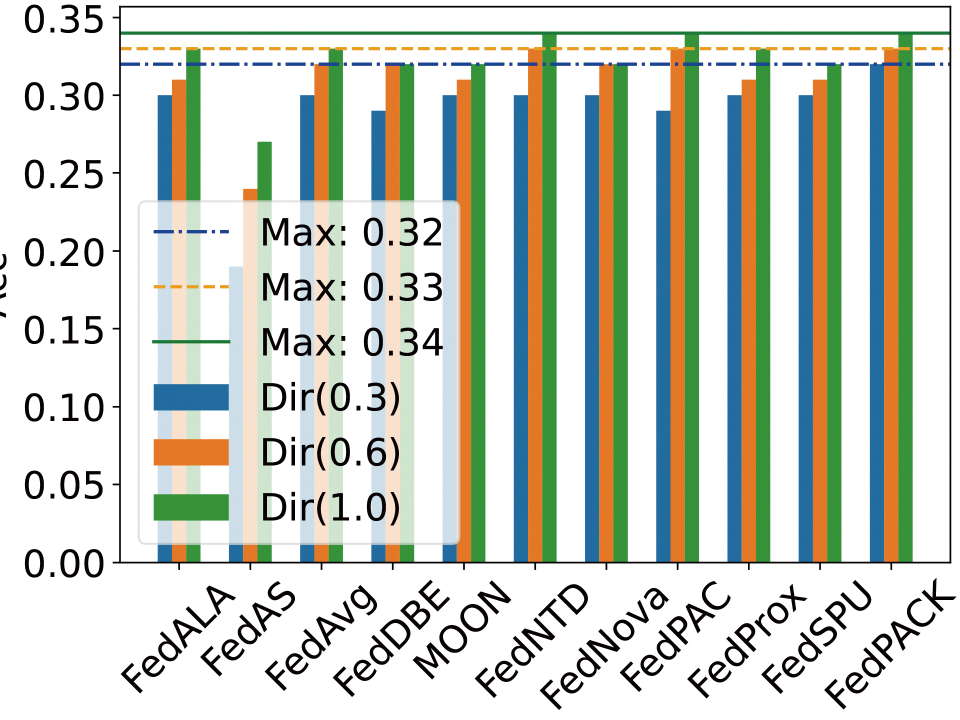}
    \label{fig4-cpr3}
  }
    \subfloat[CPR=0.3,Pathological]{
    \includegraphics[height=0.18\linewidth,width=.235\linewidth]{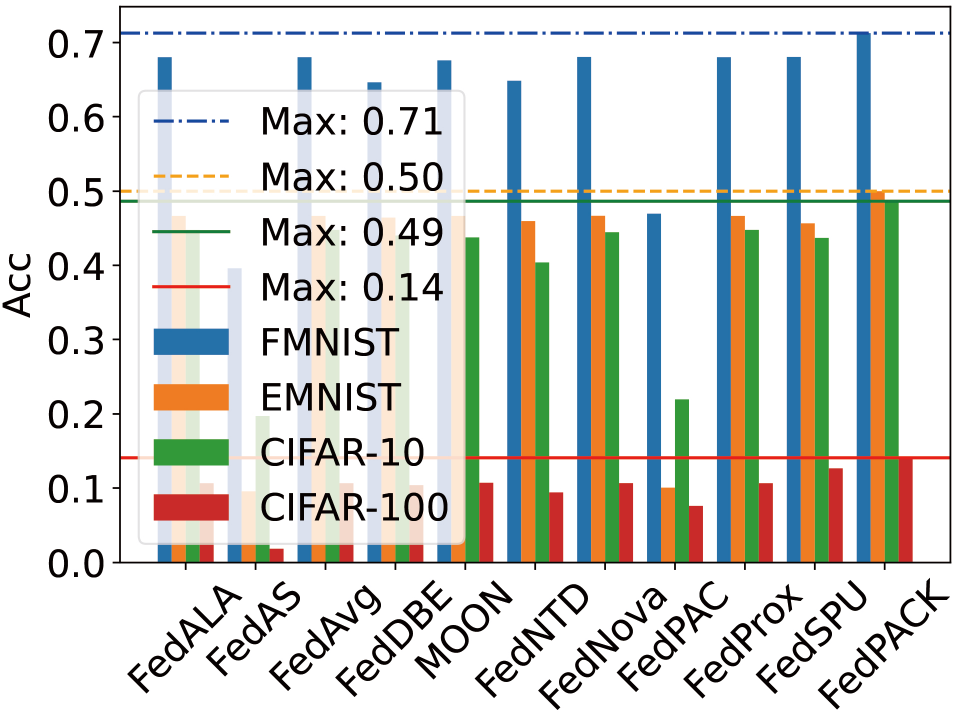}
    \label{fig4-cpr4}
  }
  \caption{The influence of the limited resources client participation ratio (CPR) on CIFAR-100, where CPR represents the ratio of the number of clients per round to the total number of clients. The horizontal line indicates the accuracy of the FedCSPACK.}
  \label{figure:fig4}
\end{figure*}

\subsection{Resource Consumption}
Table~\ref{table:t2} presents comparative results demonstrating improved computational performance and optimal resource usage. 
\textbf{Communication Traffic}. FedCSPACK significantly reduces transmission overhead across various datasets. For example, on EMNIST, the communication traffic dropped from 18.18GB to 0.73GB, achieving a compression ratio of 96.0\%, the lowest among all compared methods. On CIFAR-100, FedCSPACK compressed the communication traffic of the ResNet18 from 251.00GB to 9.24GB, representing a 27$\times$ reduction; it is 5$\times$ lower than FedSPU. This result demonstrates that FedCSPACK's packet-level sparse communication mechanism effectively reduces the number of uploaded parameters, demonstrating exceptional compression capabilities, particularly for a complex model, significantly outperforming SOTA methods. 

\noindent\textbf{Training Time}. On EMNIST and CIFAR-10, FedCSPACK's training time barely increases compared to the baseline, remaining highly stable. On CIFAR-100, although training time increases slightly from 0.81H to 0.88H, FedCSPACK is still 89.8\%, 47.7\%, and 38.11\% lower than Split, Distillation, and FedSPU, respectively, demonstrating superior computational efficiency. These results demonstrate that while the Top-K parameter pack selection mechanism introduces a small amount of additional computation, the resulting increase in training time is minimal, and even improves it in some scenarios.

\subsection{Numbers of Limited Resources Clients}
Fig.~\ref{figure:fig4} shows the impact of CPR on the model's performance and stability. In figs~\ref{fig4-cpr1} to~\ref{fig4-cpr3}, model performance fluctuates significantly as client participation declines. In low-participation scenarios (fig.~\ref{fig4-cpr3}), the impact of data heterogeneity on existing SOTA methods is further exacerbated due to the incomplete distribution of available data, leading to a significant performance drop. However, FedCSPACK maintains good stability and robustness, achieving an accuracy of 0.32, 0.33, and 0.34. In fig.~\ref{fig4-cpr4}, FedCSPACK demonstrates superior model performance in four different pathological data partitioning scenarios. This result further demonstrates FedCSPACK's effectiveness and generalization capabilities under diverse data distributions. More participation rate results are in Appendix D.

\subsection{Ablation Studies}
\textbf{Influence of Weight.} We assessed the training impact of key weights: cosine similarity (CS) weight and KL divergence weight on model performance in FedCSPACK to demonstrate their individual effectiveness. As shown in table~\ref{table:t3}, compared to using CS alone, FedCSPACK using dual-weighted aggregation improved the average accuracy by 0.01, 0.06, 0.04, and 0.05, respectively; and compared to using KL alone, it improved the accuracy by 0.03, 0.06, 0.03, 0.07, and 0.1, respectively. The result demonstrates that the dual-weighted aggregation combining parameter direction and parameter distance can more comprehensively evaluate the contribution of the parameter package and effectively improve the training efficiency of the model.
\begin{table}[ht]
\centering
\begin{tabular}{l ccccc}
\toprule
\multirow{2}{*}{Weight}& \multicolumn{5}{c}{Round} \\
\cmidrule{2-6}
    & 10 & 30 & 50 & 70 & 100 \\
\hline				  
CS        & 0.33 & 0.62 & 0.65 & 0.70 & 0.74 \\			  
KL        & 0.30 & 0.57 & 0.68 & 0.67 & 0.69 \\
{\bf Ours} & 0.33 & 0.63 & 0.71 & 0.74 & 0.79 \\
\bottomrule
\end{tabular}
\caption{Impact of key weights on model accuracy at different training stages on CIFAR-10 and Dir(0.5).}
\label{table:t3}
\end{table}
\begin{figure}[ht]
  \centering
  \subfloat[CIFAR10 ACC]{
    \includegraphics[height=0.34\linewidth, width=.481\linewidth]{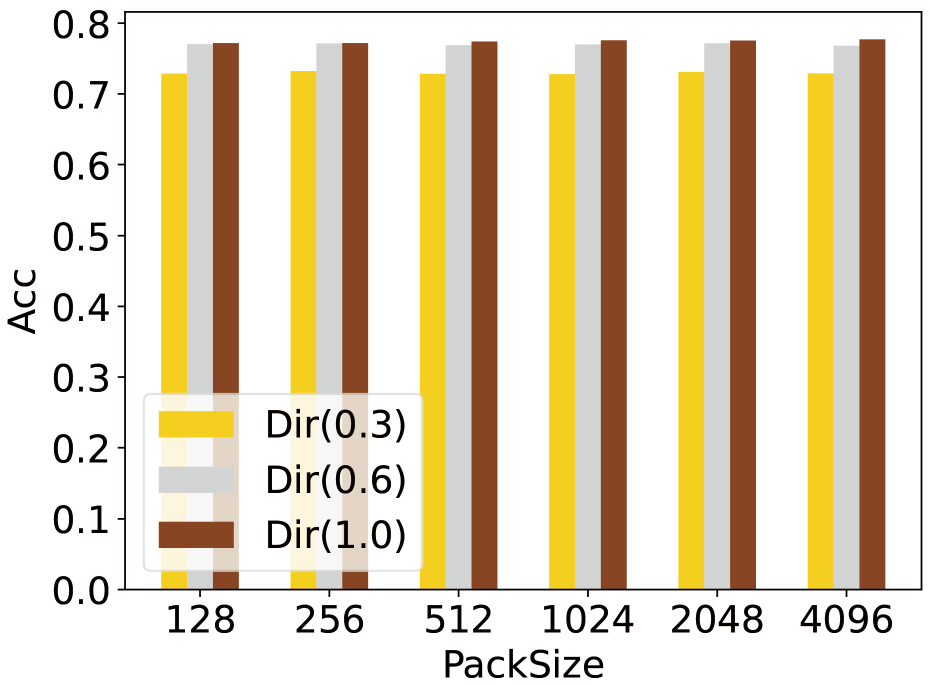}
    \label{fig5-pack1}
  }
  \subfloat[CIFAR10 Times]{
    \includegraphics[height=0.34\linewidth, width=.481\linewidth]{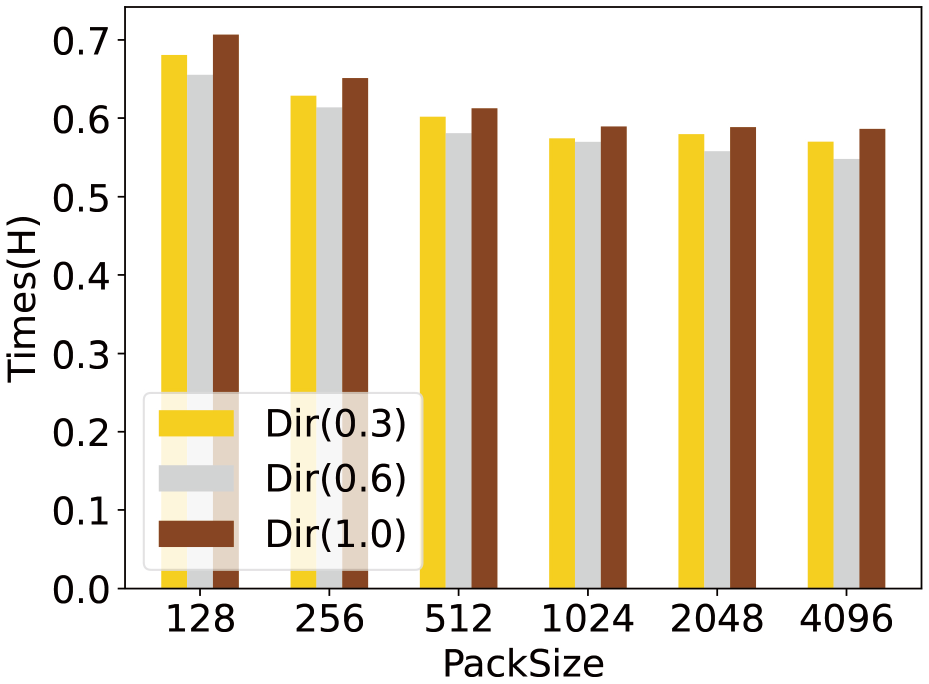}
    \label{fig5-pack2}
  }
  \\
    \subfloat[EMNIST ACC]{
    \includegraphics[height=0.34\linewidth, width=.481\linewidth]{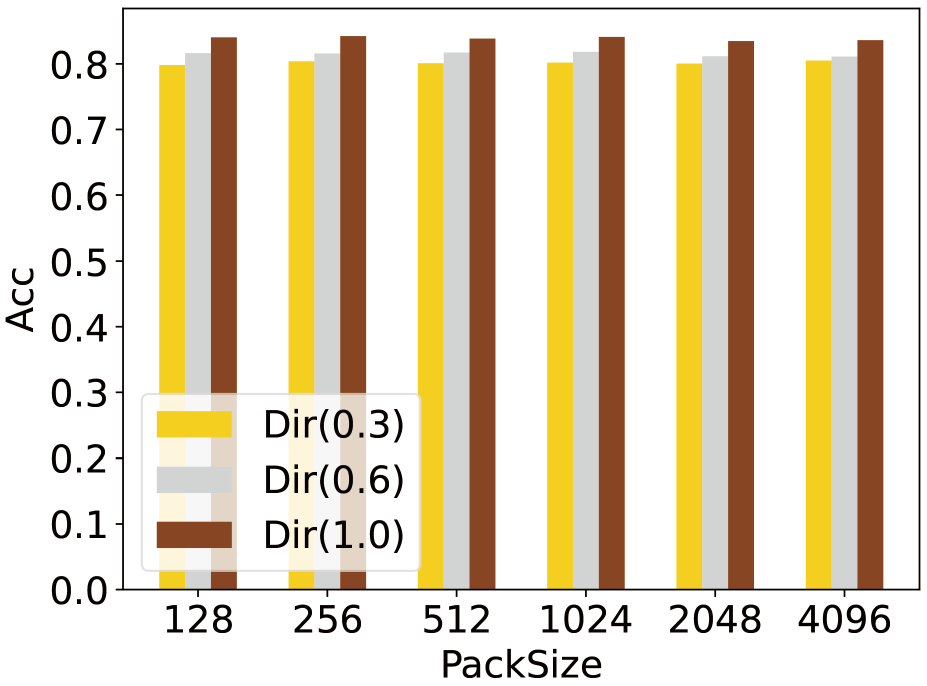}
    \label{fig5-pack3}
  }
  \subfloat[EMNIST Times]{
    \includegraphics[height=0.34\linewidth, width=.481\linewidth]{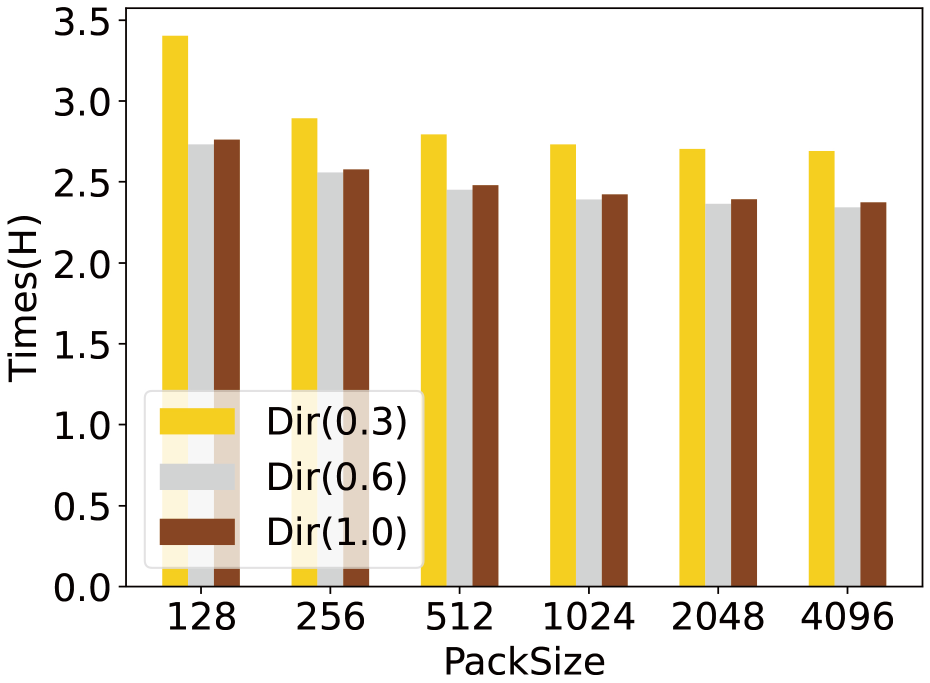}
    \label{fig5-pack4}
  }
\caption{The impact of the $PACK$ size on model performance and computation time.}
  \label{figure:fig5}
\end{figure}

\textbf{Influence of PACK.} Fig.~\ref{figure:fig5} further verifies the effect of $PACK$ on model performance. In figs~\ref{fig5-pack1} and~\ref{fig5-pack3}, as the $PACK$ increases, the overall model performance remains largely unchanged under varying data heterogeneity conditions. This demonstrates that the parameter pack selection technique selects the K most contributing local model parameter packs. The dual-weight aggregation mechanism further enhances the global model's ability to integrate local model knowledge, thereby ensuring model stability. In figs~\ref{fig5-pack2} and~\ref{fig5-pack4}, the overall time consumption of the model gradually decreases as the size of the $PACK$ increases, and the improvement of FedCSPACK is more obvious in heterogeneous scenarios with a large amount of data. Appendix E has more details.

\section{Conclusion}
In this work, we propose a novel parameter package-level PFL method (FedCSPACK) to address the coexistence of resource constraints and data heterogeneity. The key insight is to package the client model parameters, using cosine similarity and Top-k to calculate the shared parameter package, reducing the client's demand for limited resources. Through mask matrix and dual weights, FedCSPACK can ensure that the server efficiently aggregates the sparse parameter packages of client models, guaranteeing the robustness and generalization ability of the global model. Extensive experiments on various datasets have demonstrated that FedCSPACK achieves SOTA performance.

\section*{Acknowledgments}
This work is supported by the National Natural Science Foundation of China (U22B2026, 62572121, 62372231), Jiangsu Province Frontier Technology Research and Development Program (BF2025067), Aeronautical Science Foundation (2022Z071052008), Fundamental Research Funds for the Central Universities (2242025K30025), and the Big Data Computing Center of Southeast University.

\bibliography{aaai2026}

\newpage
\appendix
\setcounter{table}{0}
\setcounter{figure}{0}
\setcounter{equation}{0}
\renewcommand{\thetable}{A\arabic{table}}
\renewcommand{\thefigure}{A\arabic{figure}}
\renewcommand{\theequation}{A\arabic{equation}}
\renewcommand{\thesection}{Appendix \Alph{subsection}}

\section*{Appendix A}
\noindent\textbf{Dataset Setup}: In real-world federated learning, user data naturally diverges in distribution because of heterogeneous preferences, lifestyles, and the distinct spatial–temporal contexts. This divergence manifests as a non-independent and identically distributed (Non-IID) challenge on every client, leading to degraded model accuracy, slower convergence speed, and lower model availability. Suppose a federated learning task is carried out on feature x and label y. Users need to extract sample data from their private data to complete local training, that is, $(x,y)\sim p(x,y)$, which can be split into $p(x,y) = p(x|y)p(y)$. Since estimating $p(x|y)$ is costly, while estimating the label prior $p(y)$ is extremely lightweight, we take the imbalance of label distribution as an efficient entry point for Non-IID partitioning - the core idea is to have significantly different label distributions for different clients. Based on this, we adopt the two most classic partitioning strategies in the field of federated learning: \textbf{Dirichlet Sampling} and \textbf{Pathological Sampling}. 
\begin{figure}[htbp]
    \centering
    \subfloat[$Dir(1.0)$\\Client $\to$ Lable]{
        \includegraphics[width=.48\linewidth]{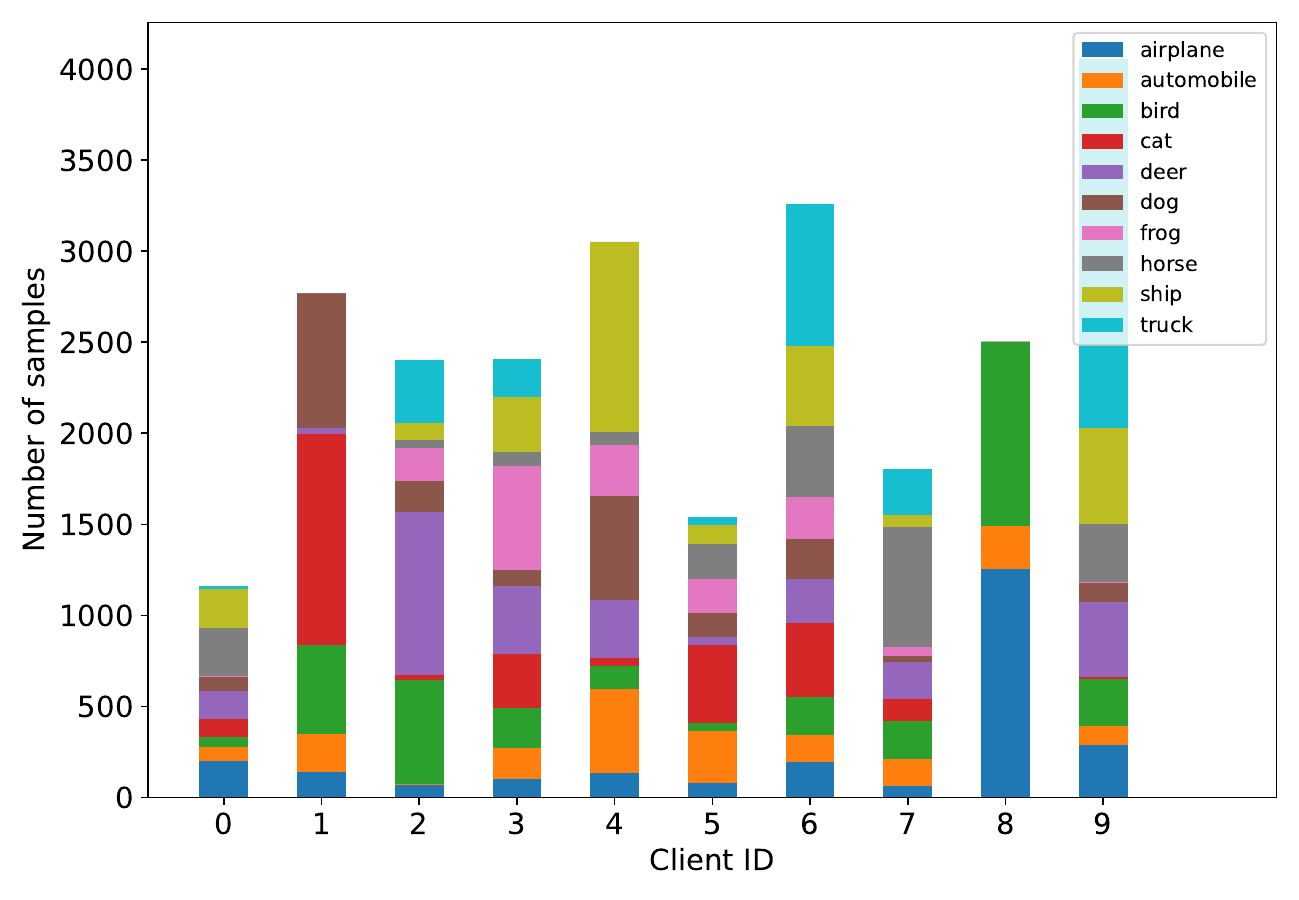}
        \label{app-fig1:cl1}
    }
    \subfloat[$Dir(1.0)$\\label $\to$ client]{
        \includegraphics[width=.48\linewidth]{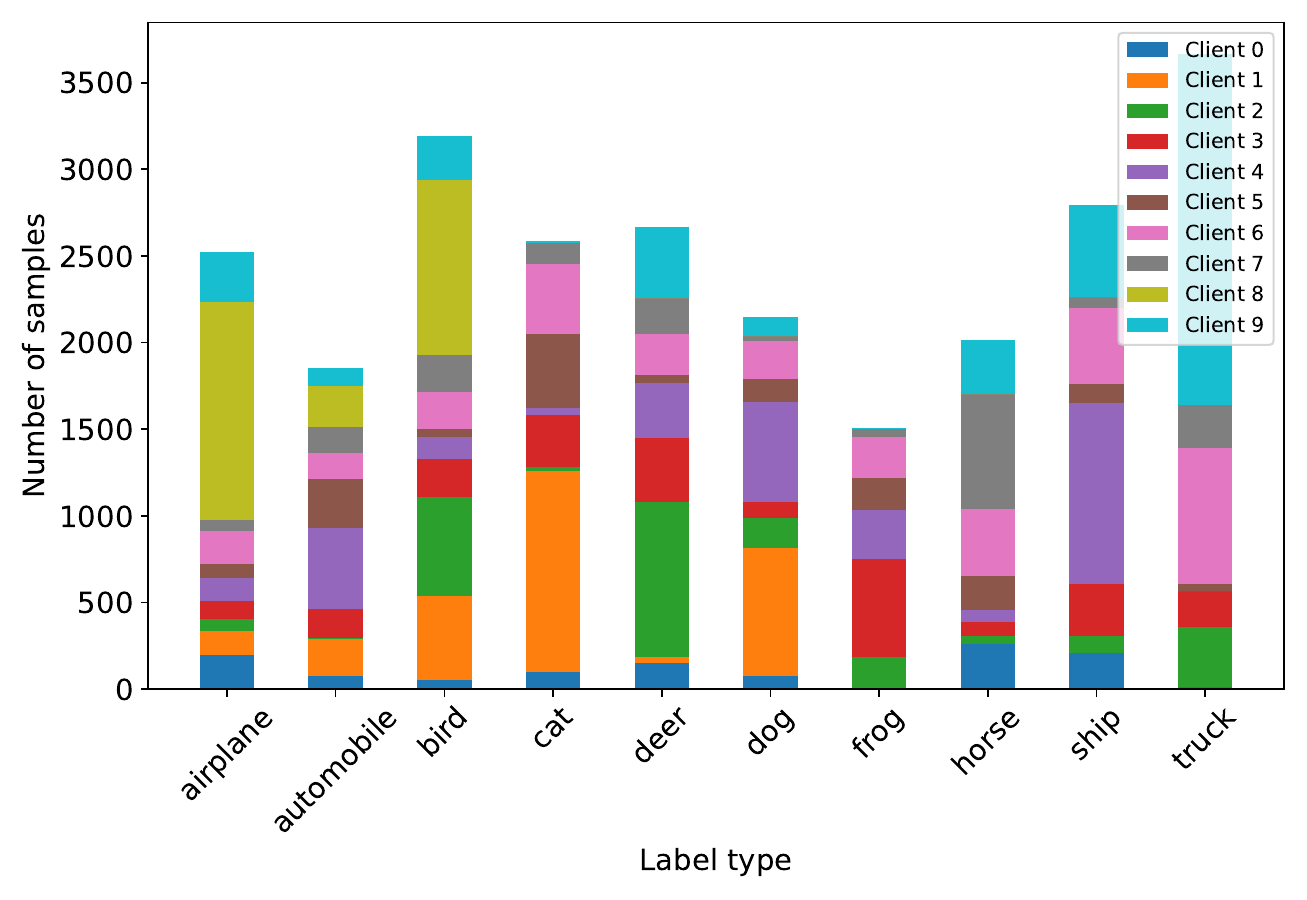}
        \label{app-fig1: cl2}
    }
    \\
    \subfloat[$Pathological$\\Client $\to$ Lable]{
        \includegraphics[width=.48\linewidth]{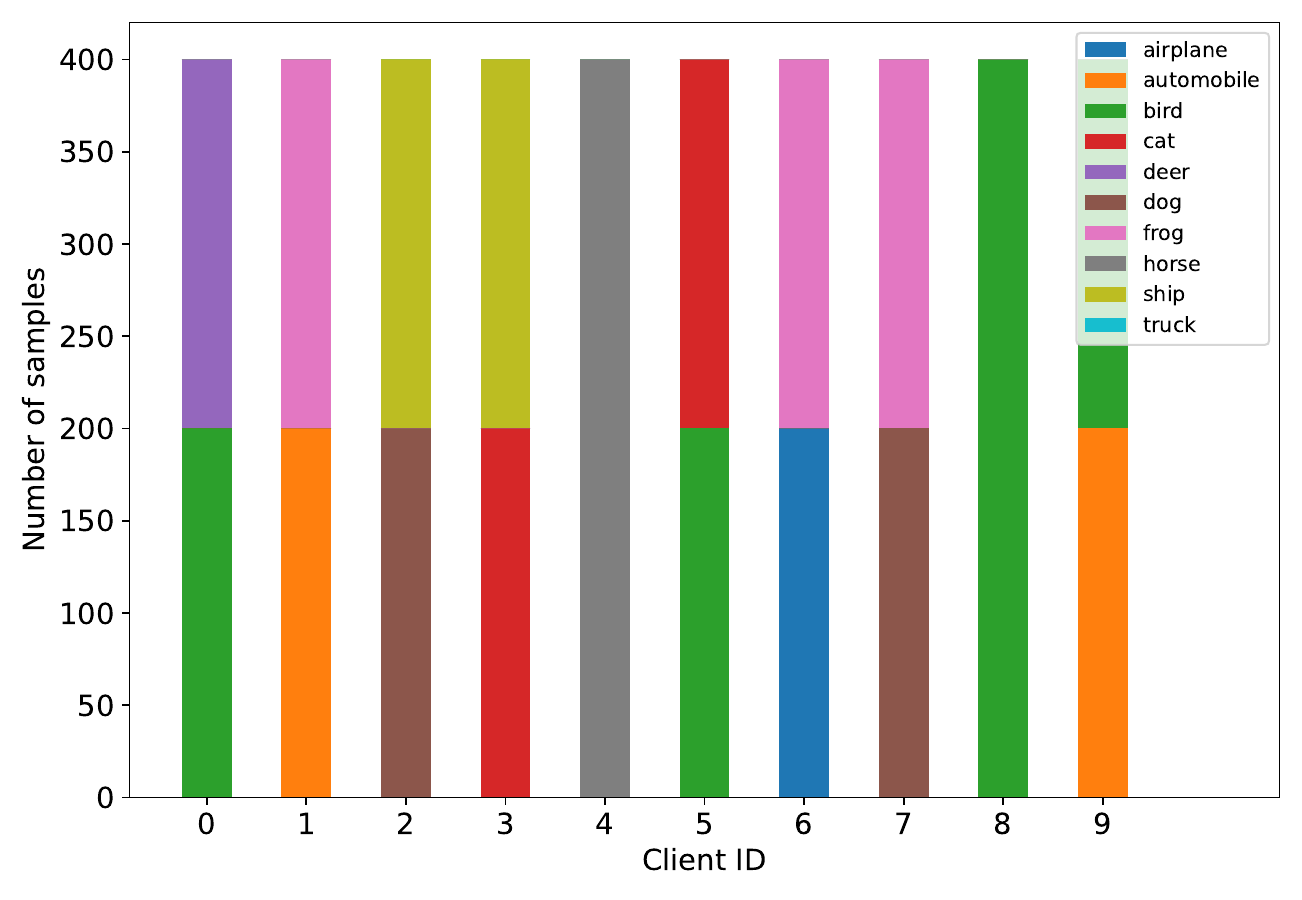}
        \label{app-fig1: cl3}
    }
    \subfloat[$Pathological$\\label $\to$ client]{
        \includegraphics[width=.48\linewidth]{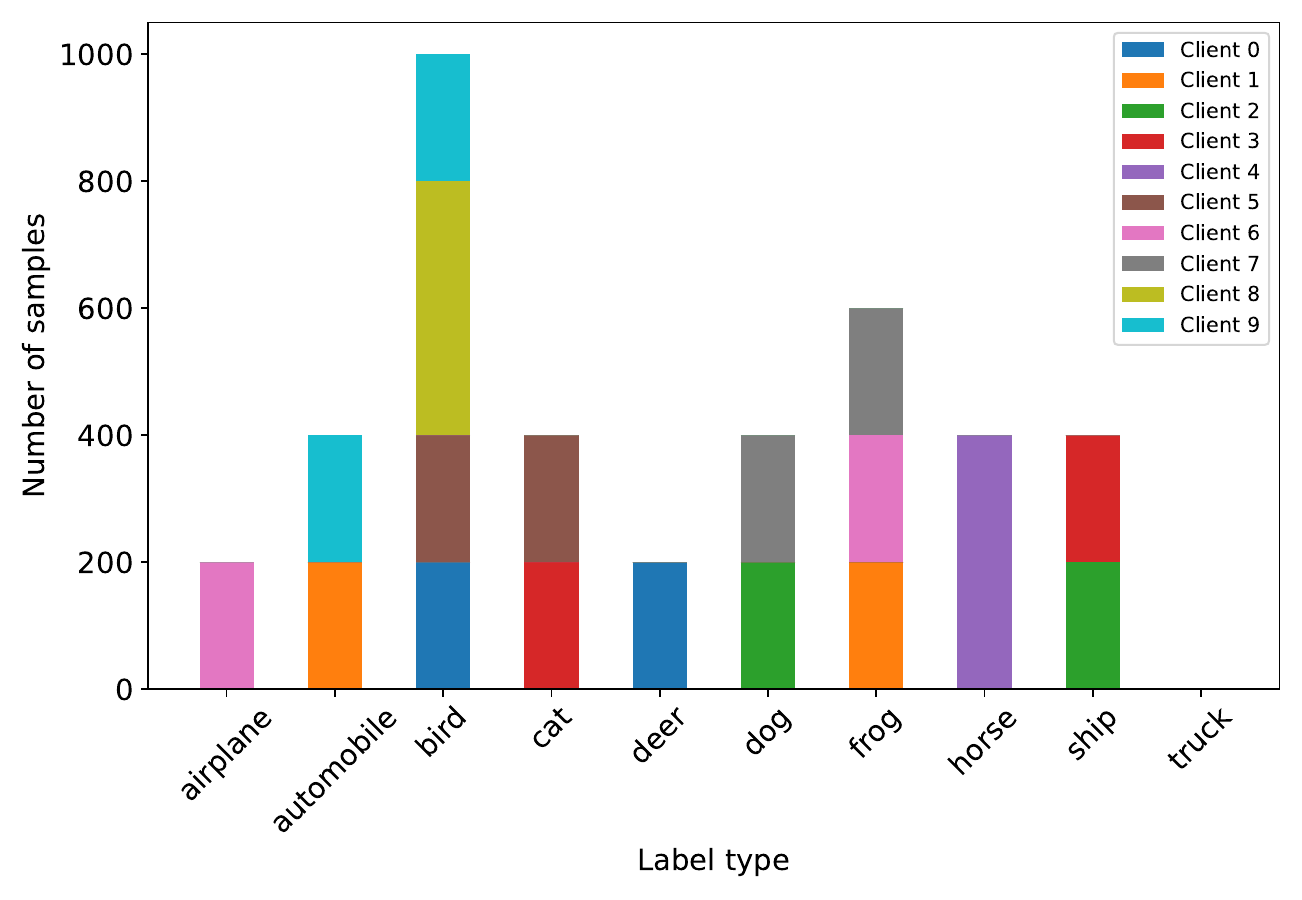}
        \label{app-fig1: cl4}
    }
    \caption{Data heterogeneity of 10 clients was simulated using Dirichlet and shard on CIFAR-10.}
    \label{app-fig1}
\end{figure}

In \textbf{Dirichlet Sampling}, the distribution pattern is regulated by the concentration parameter $\alpha$. Fig.~\ref{app-fig1:cl1} visually demonstrates this effect: the smaller the $\alpha$, the sharper the label distribution, and the stronger the heterogeneity among clients; Conversely, they tend to converge. In contrast to Dirichlet's "probabilistic heterogeneity", \textbf{Pathological Sampling} adopts ``physical heterogeneity'': first, the 10 categories of CIFAR-10 are equally divided into several data blocks, and then randomly and non-overlap them are assigned to 10 clients. Fig.~\ref{app-fig1: cl3} shows that each client ultimately only has a few complete category data blocks. This ``category isolation'' is closer to real-world scenarios such as medical institutions and mobile terminals that naturally collect data by business/region, thereby providing another verification benchmark for strong Non-IID. Overall, these two methods jointly cover Non-IID scenarios ranging from mild to severe, providing a standard test benchmark for the robustness evaluation of federated learning algorithms.

\noindent\textbf{Parameter Settings and System Implementation.}: The maximum global iteration is set to T = 100 with a total of clients (M =100). The number of active clients per round is set to 10, and each client has three/five local training epochs. The learning rate is set to $0.001-0.05$, and $0.05-0.1$ respectively for FMNIST, EMNIST, CIFAR-10, and CIFAR-100. The batch size is set to 16/32/64/128 for FMNIST and EMNIST, and 16/32 for CIFAR-10/CIFAR-100. To validate the effectiveness of FedCSPACK, we used three CNN models (each containing 3 convolutional layers and 3 linear layers) and a ResNet18 model as the base model, with all methods using SGD as the local optimizer. The experiment is implemented with Pytorch 2.0.0 and NVIDIA 3090 Ti GPU in a native Linux environment, and each method is run three times to report the average result.

\section*{Appendix B}
\textbf{The overall performance of the global model}. 

\noindent We experimentally evaluate all the schemes in three heterogeneous data environments, and the specific results are presented in Table. \ref{app-table:t1}. As shown in Table \ref{app-table:t1}, FedCSPACK outperforms all comparative methods across all evaluated datasets. It achieved the highest performance metrics on FMNIST, EMNIST, CIFAR-10, and CIFAR-100, and is far ahead of other solutions. On FMNIST, FedCSPACK reaches 70.28\%, outperforming the second-ranked FedNova (68.06\%) by approximately 2.2\%. On EMNIST, its score of 49.00\% surpasses the next-best methods (FedAvg and FedALA, both at 46.67\%) by around 2.3\%. 

In CIFAR-10 subset with generally lower performance, FedCSPACK maintains a clear lead at 13.08\%, exceeding MOON (10.71\%) by roughly 2.4\%. While other methods (e.g., FedAS, FedPAC) exhibit severe underperformance on specific datasets (e.g., FedAS scores only 9.55\% on EMNIST), FedCSPACK delivers consistently leading results across diverse datasets. In contrast to traditional methods like FedAvg and FedProx, which show stagnant performance across datasets (e.g., 68.03\% on FMNIST), FedCSPACK achieves systematic breakthroughs—with its advantages being particularly pronounced on more complex datasets such as CIFAR-10 and EMNIST. In summary, through dual-weight aggregation, FedCSPACK maintains high model accuracy, demonstrating effective adaptability to datasets spanning from simple to complex and from small-scale to large-scale.
\begin{table*}[htbp]
\centering
\small
\renewcommand{\arraystretch}{1.2}
\resizebox{1\textwidth}{!}{
\begin{tabular}{l cccccccccc>{\columncolor{gray!10}}c}

\toprule
Method & FedALA  & FedAS   & FedAvg  & FedDBE  & MOON    & FedNTD  & FedNova  & FedPAC  & FedProx & FedSPU & FedCSPACK \\
\hline
FMNIST    & 68.03   & 39.60   & 68.03   & 64.64   & 67.60   & 64.85   & 68.06    & 46.97   & 68.03   & 68.18 & 70.28     \\
EMNIST    & 46.67   & 9.55    & 46.67   & 46.44   & 46.67   & 45.97   & 46.68    & 10.05   & 46.67   & 44.18 & 49.00     \\
CIFAR-10  & 44.78   & 19.72   & 44.78   & 44.02   & 43.77   & 40.39   & 44.47    & 21.97   & 44.78   & 43.21 & 47.65     \\
CIFAR-100 & 10.65   & 1.87    & 10.65   & 10.39   & 10.71   & 9.41    & 10.66    & 7.59    & 10.65   & 11.13 & 13.08     \\
\bottomrule
\end{tabular}
}
\caption{The test accuracy (\%) across the Pathological Sampling on four datasets.The background color represents our method.}
\label{app-table:t1}
\end{table*}
\begin{figure*}[htbp]
    \centering
    \subfloat[Dir(0.3),FMNIST]{
        \includegraphics[height=.23\linewidth,width=.33\linewidth]{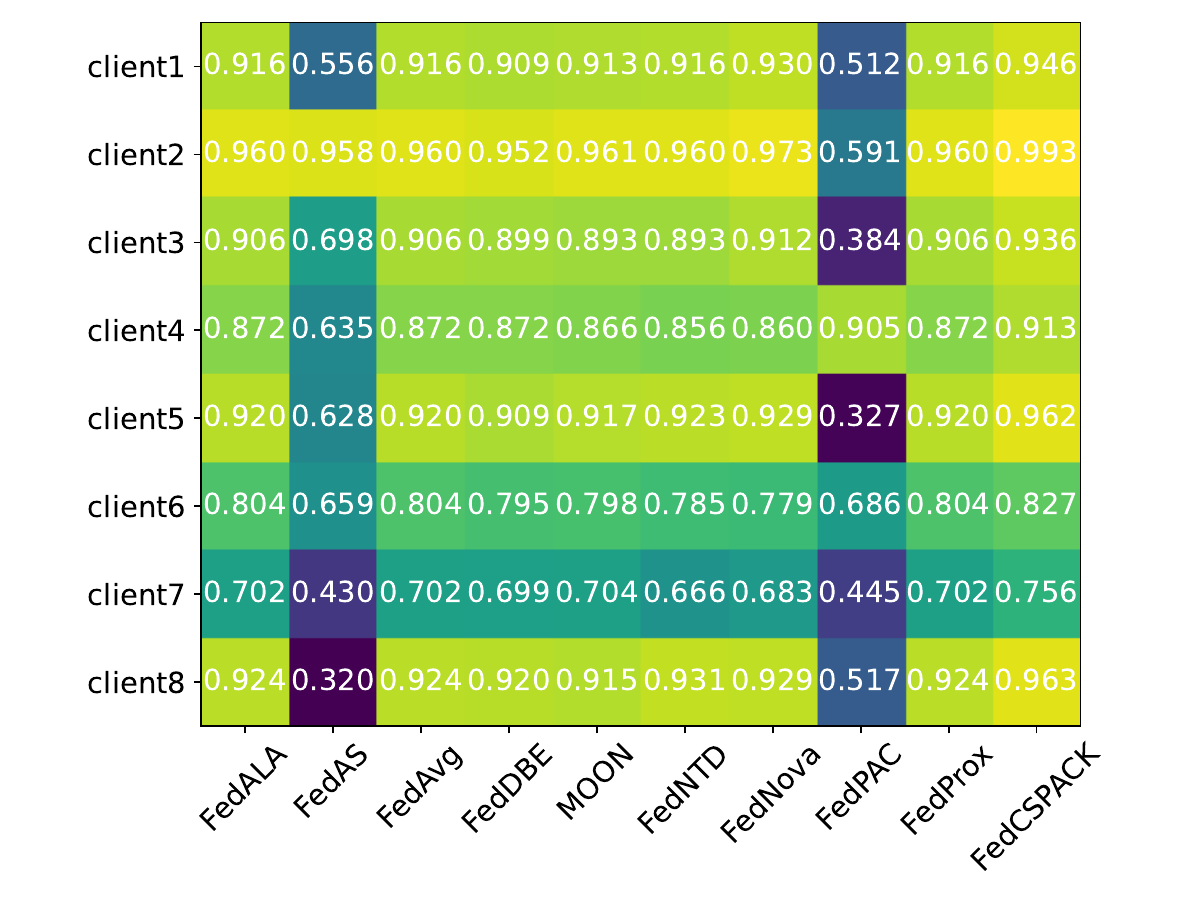}
        \label{app-fig2:cl1}
    } 
    \subfloat[Dir(0.3),EMNIST]{
        \includegraphics[height=.23\linewidth,width=.33\linewidth]{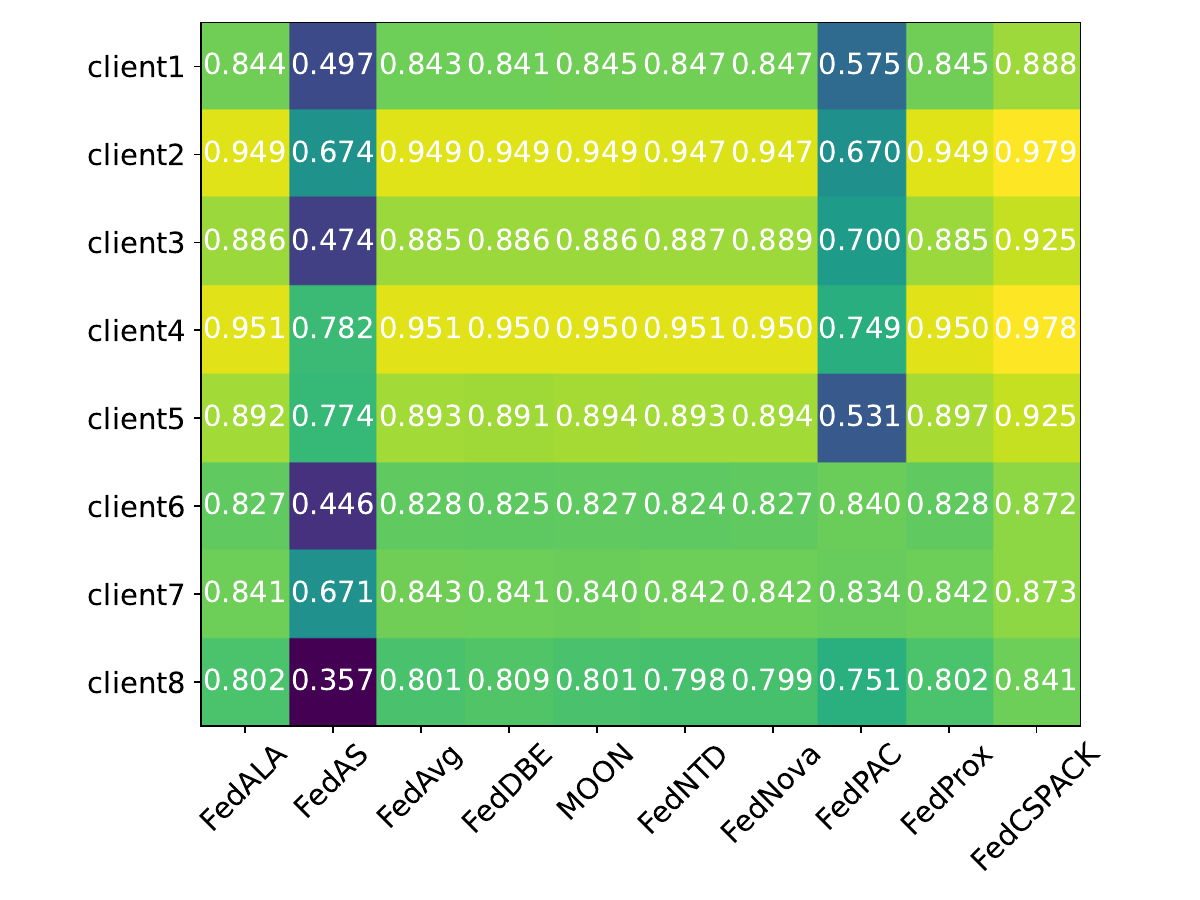}
        \label{app-fig2:cl4}
    }
    \subfloat[Dir(0.3),CIFAR-100]{
        \includegraphics[height=.23\linewidth,width=.33\linewidth]{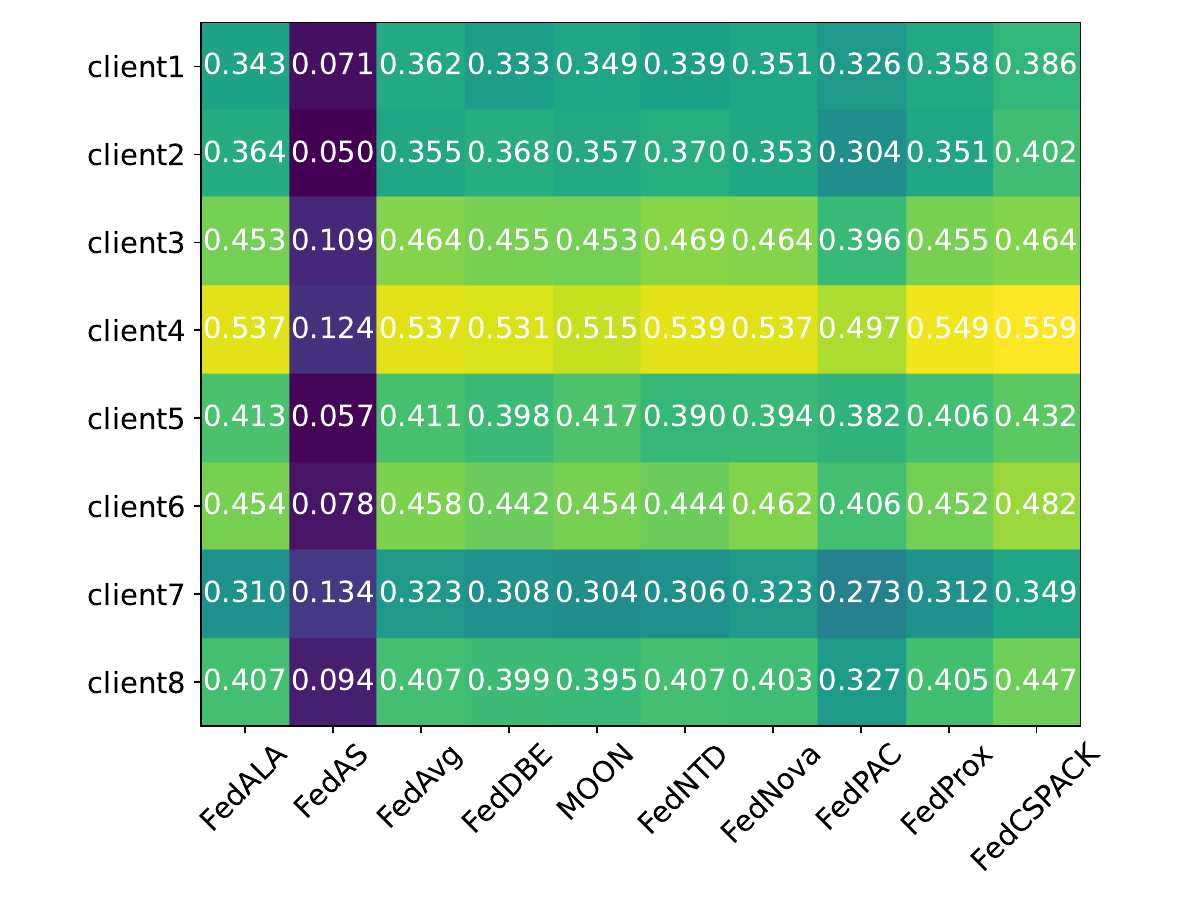}
        \label{app-fig2:cl10}
    }
	\\
    \subfloat[Pathological,FMNIST]{
        \includegraphics[height=.33\linewidth,width=.49\linewidth]{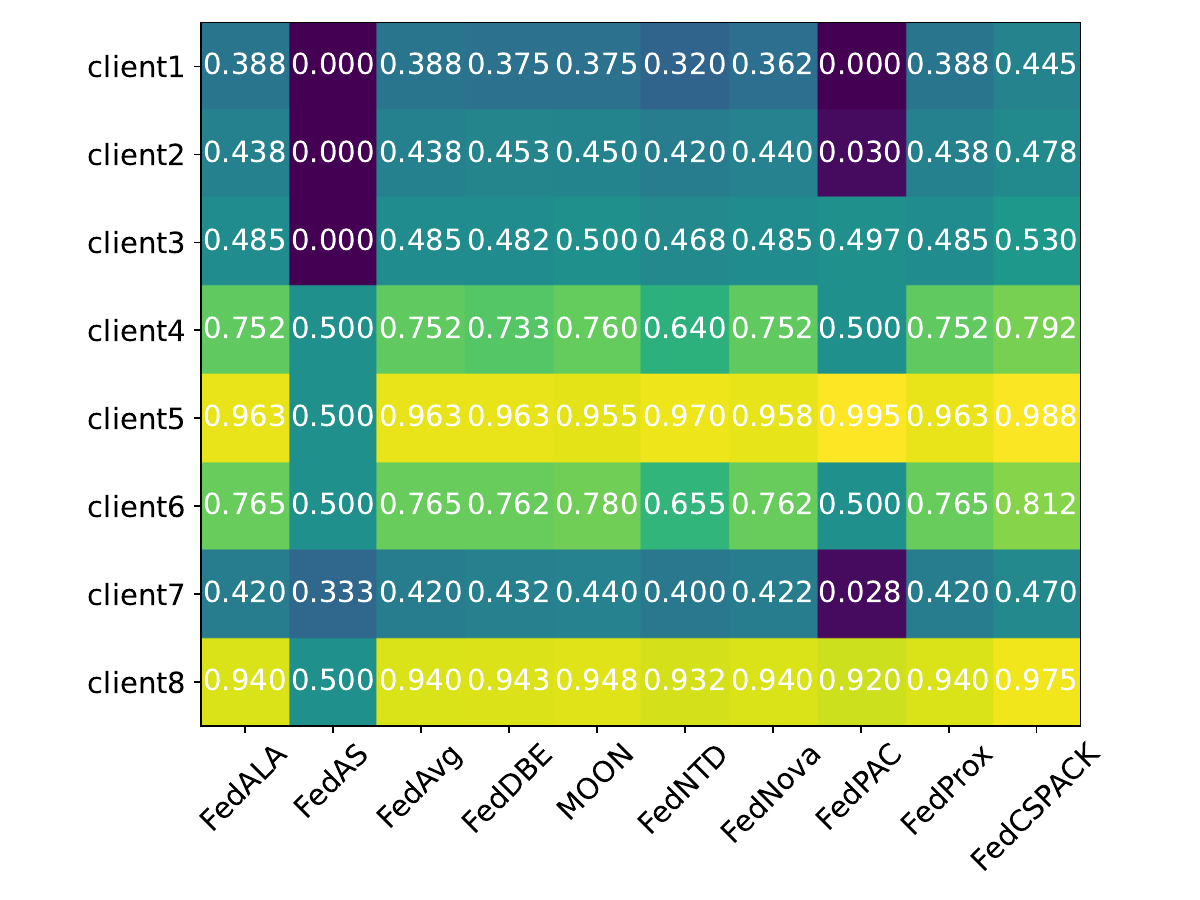}
        \label{app-fig22:cl13}
    }
    \subfloat[Pathological,EMNIST]{
        \includegraphics[height=.33\linewidth,width=.49\linewidth]{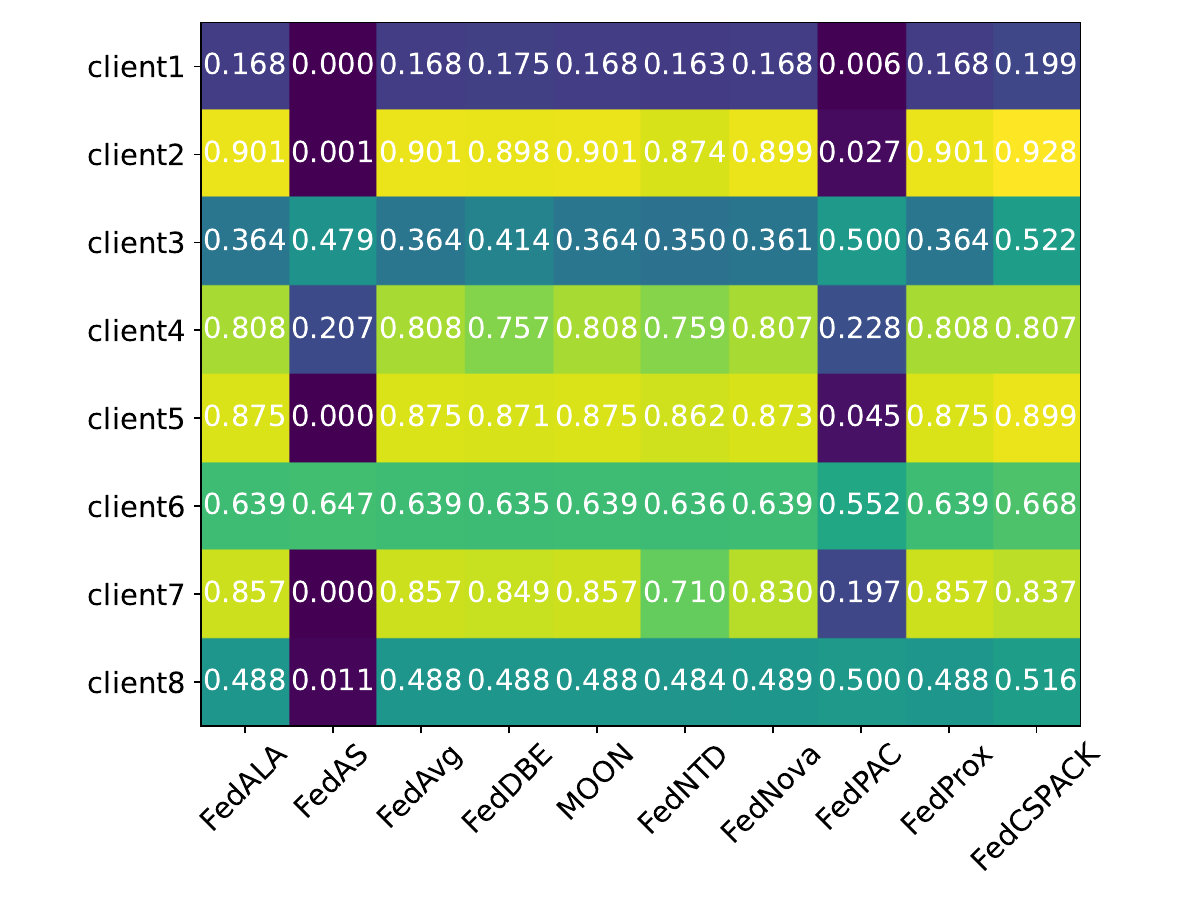}
        \label{app-fig22:cl14}
    }
    \\
    \subfloat[Pathological,CIFAR-10]{
        \includegraphics[height=.33\linewidth,width=.49\linewidth]{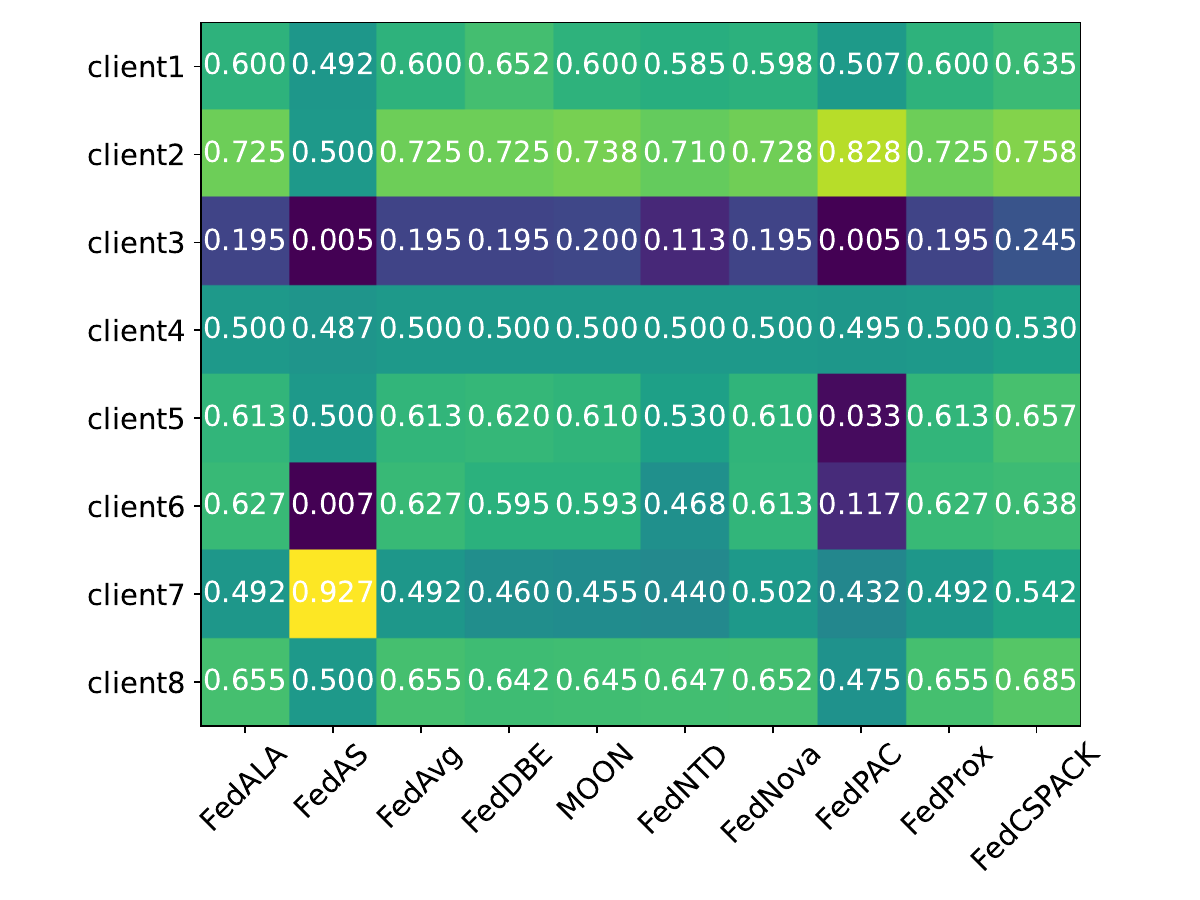}
        \label{app-fig22:cl15}
    }
    \subfloat[Pathological,CIFAR-100]{
        \includegraphics[height=.33\linewidth,width=.49\linewidth]{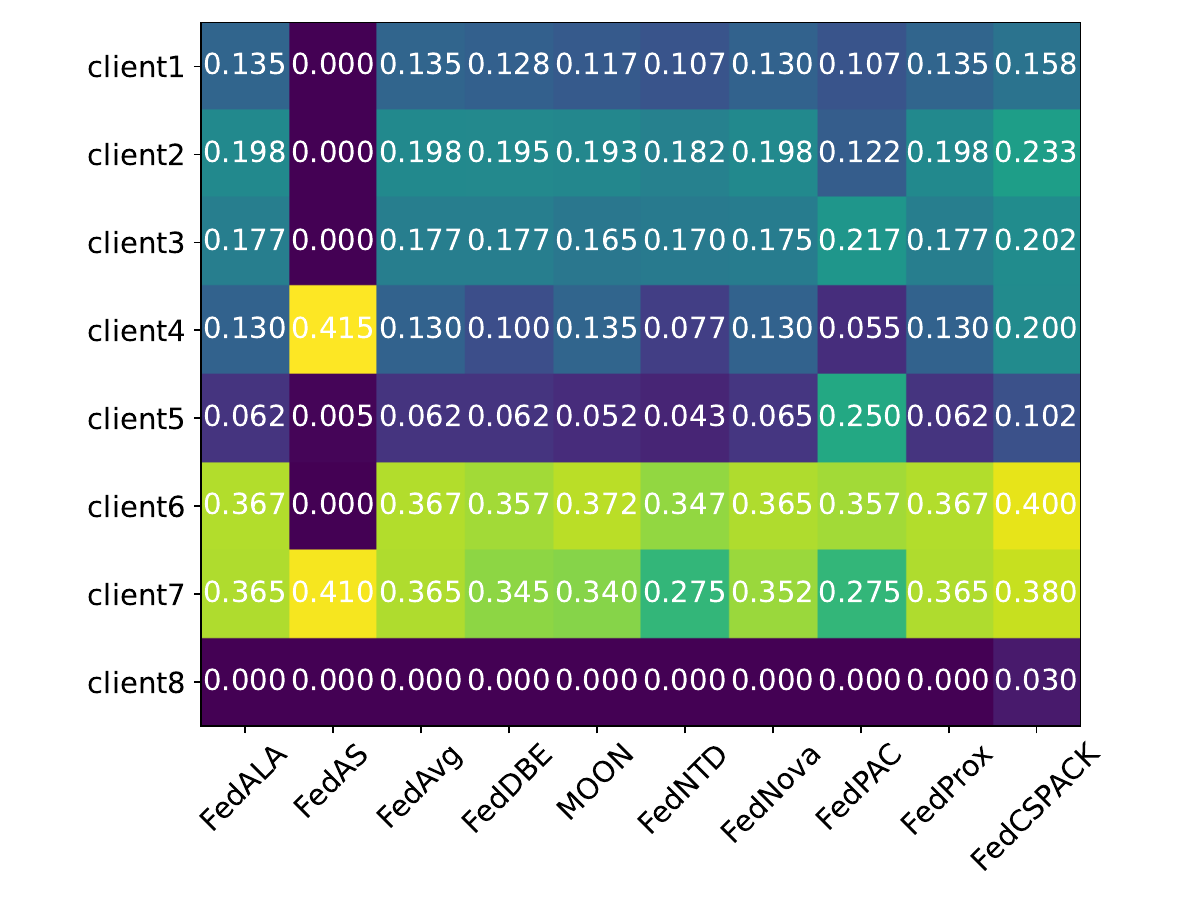}
        \label{app-fig22:cl16}
    }
    \caption{The generalization of the global model on the clients ($Dir(0.3)$ and Pathological Sampling).Use the CNN model for testing}
    \label{app-fig2}
\end{figure*}

\section*{Appendix C}
\textbf{The generalization ability of the global model on the client}. 

\noindent Fig.~\ref{app-fig2} illustrates that among the eight local clients, the global model trained by FedCSPACK shows effective model performance in adapting to the heterogeneous data of the local client. As shown in fig.~\ref{app-fig2}, FedCSPACK's global model consistently achieves significantly higher local test accuracy than other schemes across both Dirichlet and shard partitioning scenarios. When varying the Dirichlet parameter $\alpha$ from 0.3 to 1.0, all methods exhibit improved client-side generalization, but FedCSPACK demonstrates the most substantial gains under strong Non-IID conditions. This highlights its exceptional robustness against severely skewed label distributions. When $alpha = 0.3$, the local worst values of FMNIST, EMNIST, CIFAR-10, and CIFAR-100 were about 32\%, 35.7\%, 1.3\%, and 5\%, and the improvement rate was 96.3\%, 84.1\%, 76.8\%, and 40.2\% for FedCSPACK, respectively.

For fig.~\ref{app-fig22:cl13}, Fig.~\ref{app-fig22:cl14}, Fig.~\ref{app-fig22:cl15}, and Fig.~\ref{app-fig22:cl16}, The average accuracy maintained by FedCSPACK across all clients is 68.63\%, 67.2\%, 58.63\%, and 21.31\%, respectively, demonstrating that the cosine top-k parameter package technology can effectively enhance the client's knowledge absorption ability and improve the generalization ability of the global model. In summary, FedCSPACK delivers marked advantages in global-to-local generalization under both Dirichlet and shard partitions. Its gains are particularly pronounced under $\alpha < 0.6$ or in highly heterogeneous shard scenarios, proving that sparse parameter packaging and dual-weight aggregation effectively train personalized models while preserving global adaptability to data heterogeneity.

\begin{figure}[ht]
    \centering
    \subfloat[CPR=0.6]{
        \includegraphics[width=.95\linewidth]{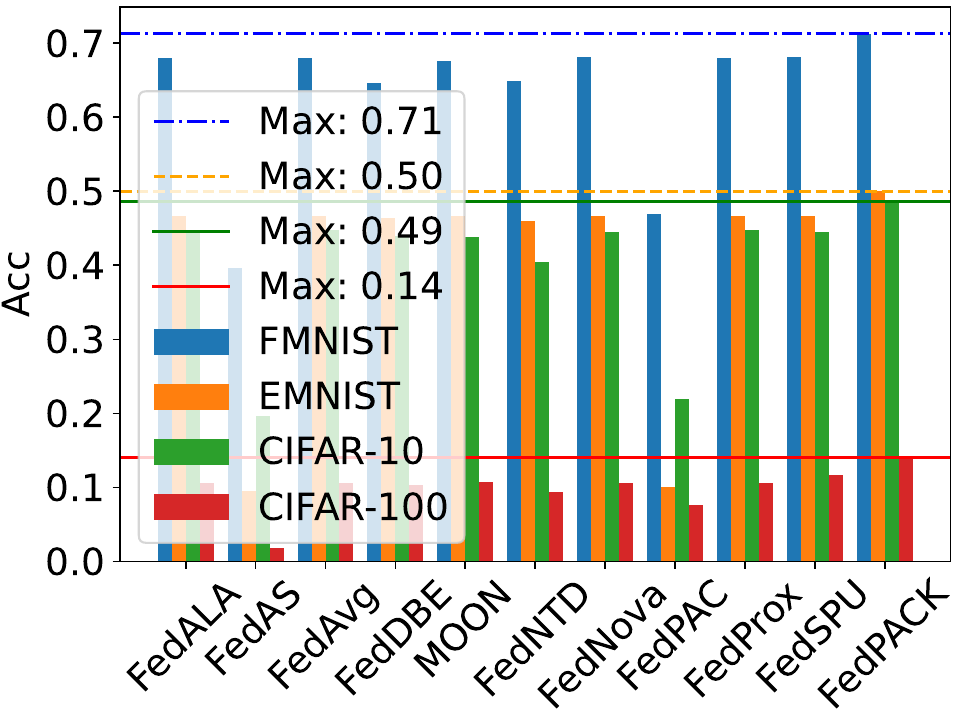}
        \label{app-fig42:c10}
    }\\
    \subfloat[CPR=1.0]{
        \includegraphics[width=.95\linewidth]{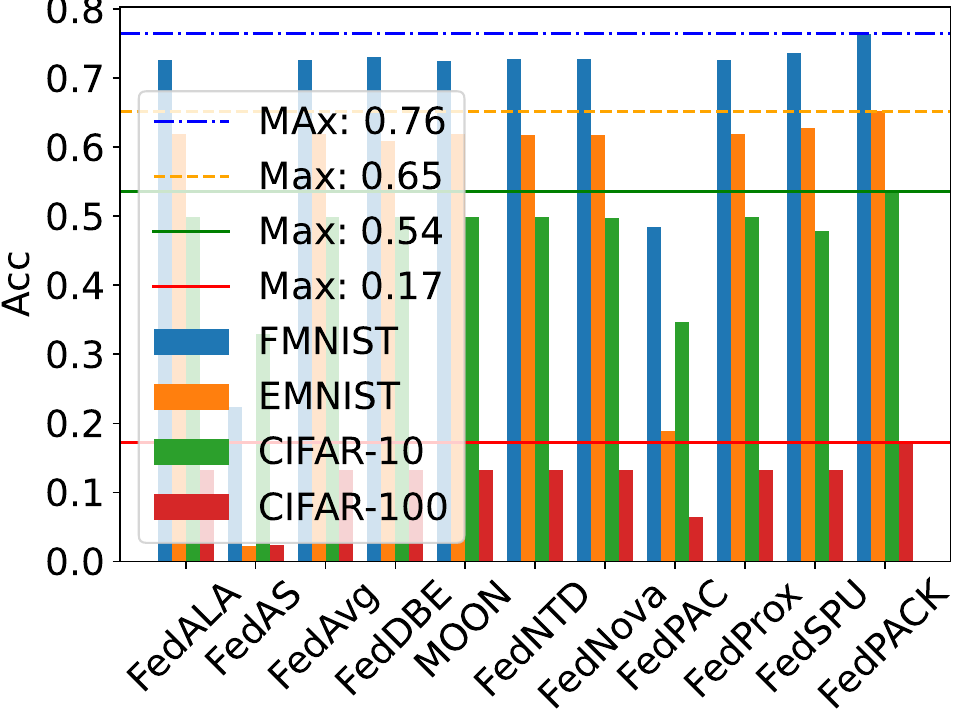}
        \label{app-fig42:c11}
    }
  \caption{The influence of the limited resources client participation ratio (CPR) on  Pathological, where CPR represents the ratio of the number of clients per round to the total number of clients. The horizontal line indicates the accuracy of the FedCSPACK. Use the CNN model for testing}
    \label{app-fig42}
\end{figure}

\begin{figure*}[htbp]
    \centering
    \subfloat[CPR=0.3, FMNIST]{
        \includegraphics[width=.33\linewidth]{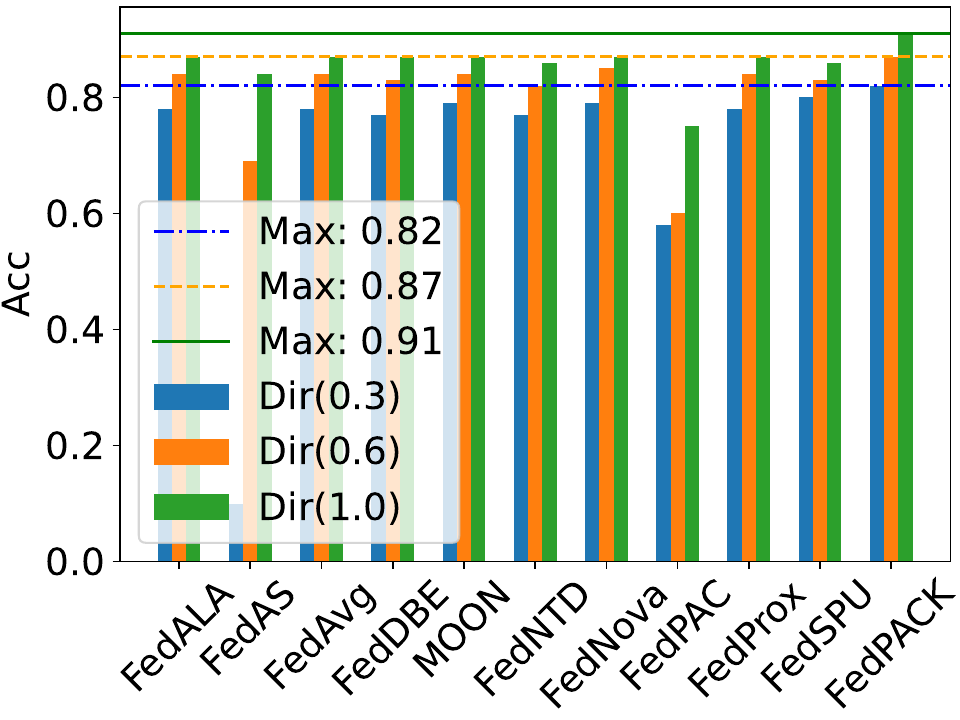}
        \label{app-fig4:c1}
    }
    \subfloat[CPR=0.6, FMNIST]{
        \includegraphics[width=.33\linewidth]{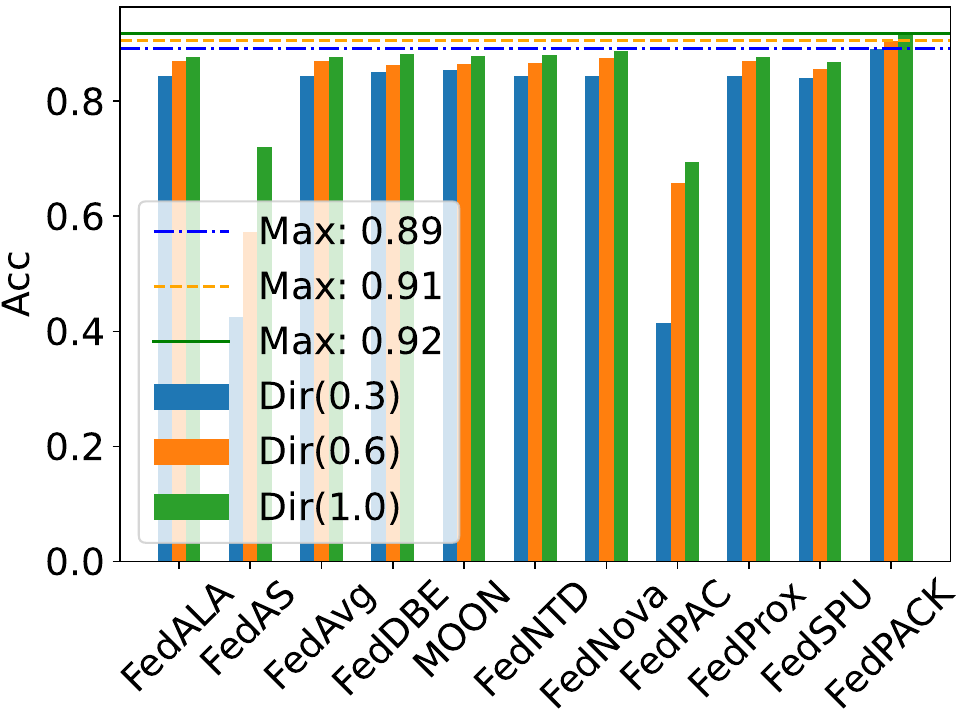}
        \label{app-fig4:c2}
    }
    \subfloat[CPR=1.0, FMNIST]{
        \includegraphics[width=.33\linewidth]{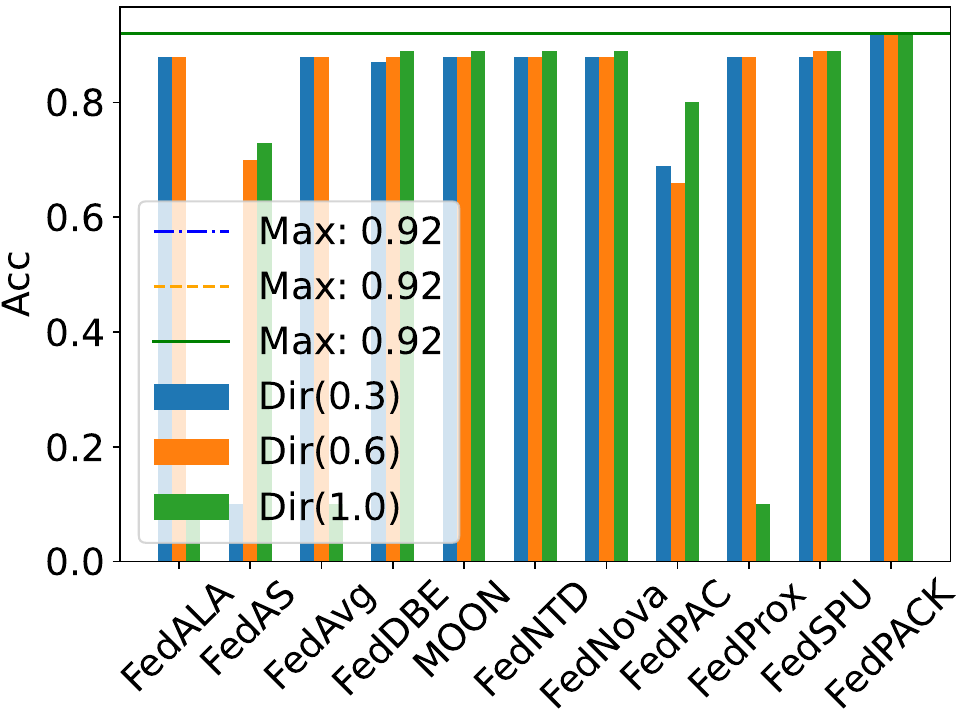}
        \label{app-fig4:c3}
    }
    \\
    \subfloat[CPR=0.3, EMNIST]{
        \includegraphics[width=.33\linewidth]{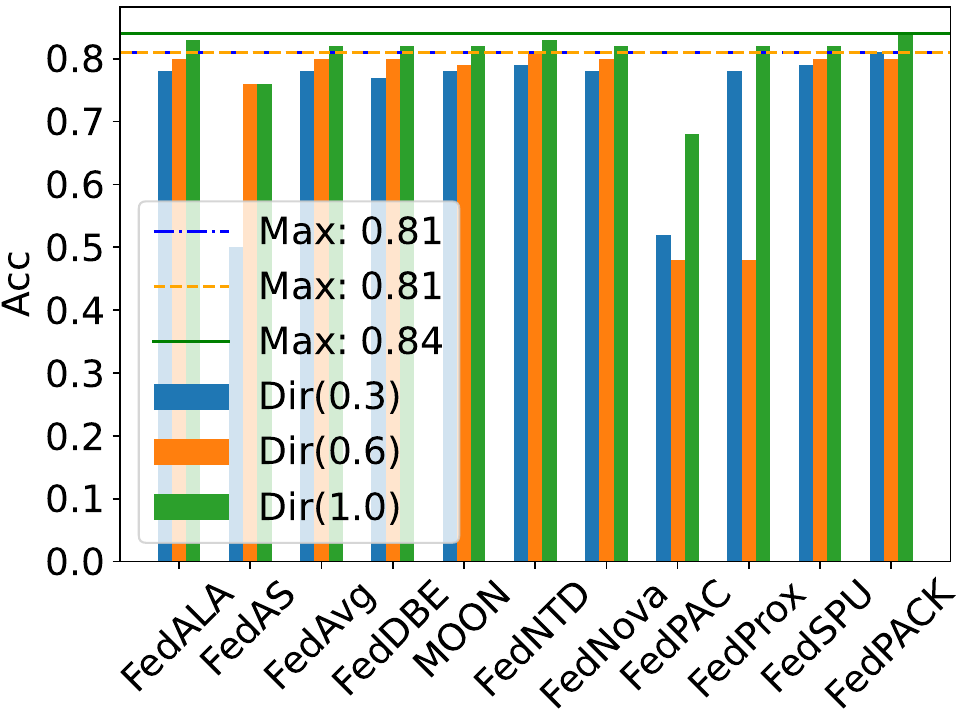}
        \label{app-fig4:c4}
    }
    \subfloat[CPR=0.6, EMNIST]{
        \includegraphics[width=.33\linewidth]{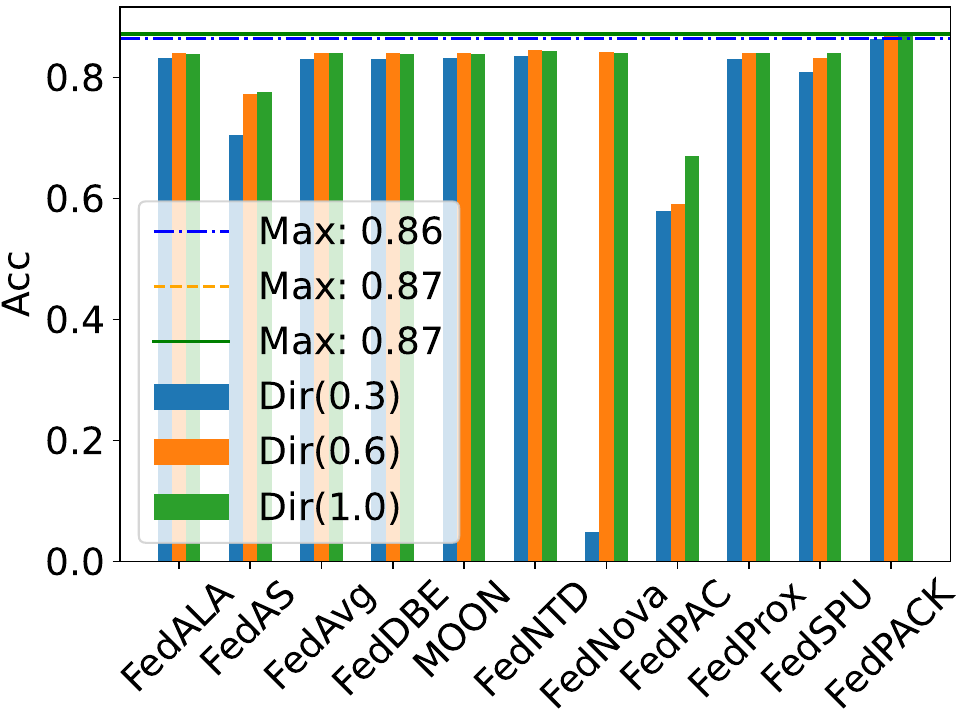}
        \label{app-fig4:c5}
    }
    \subfloat[CPR=1.0, EMNIST]{
        \includegraphics[width=.33\linewidth]{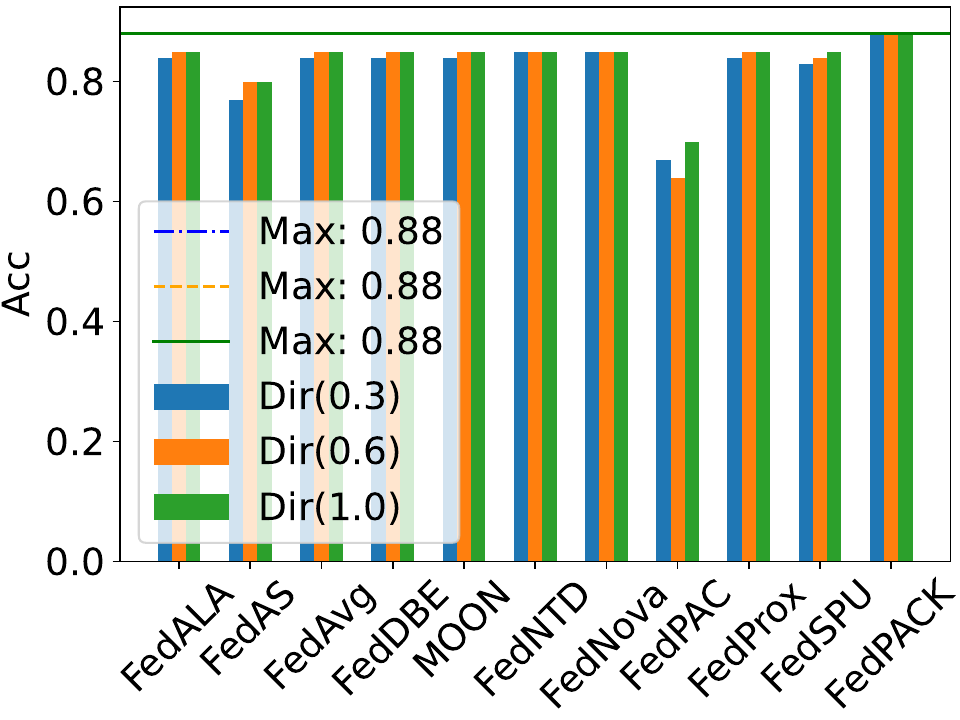}
        \label{app-fig4:c6}
    }
    \\
    \subfloat[CPR=0.3, CIFAR-10]{
        \includegraphics[width=.33\linewidth]{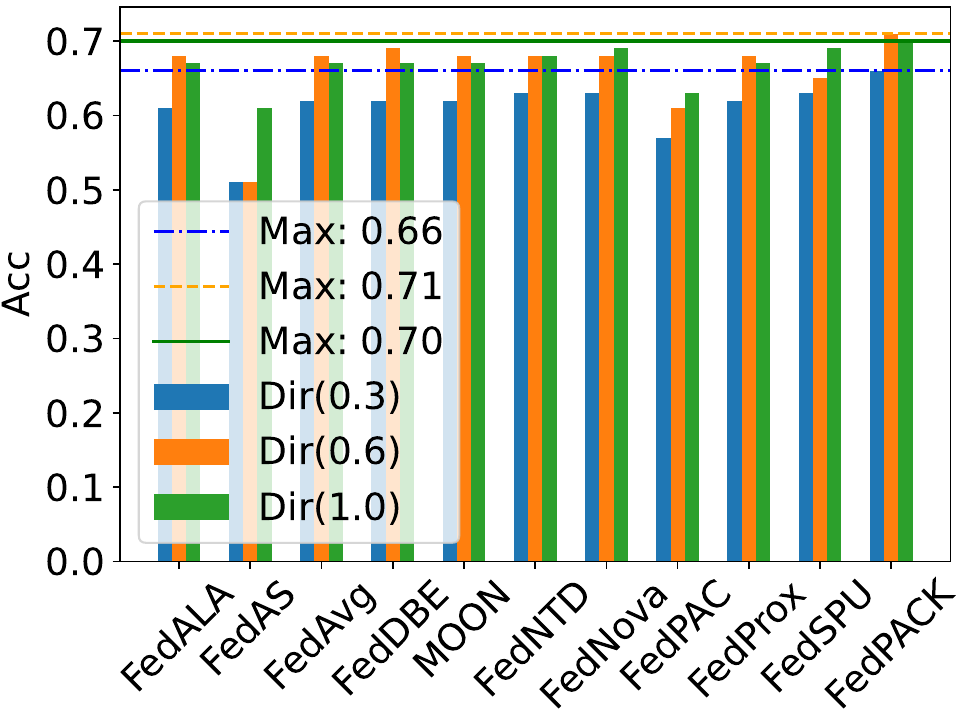}
        \label{app-fig4:c7}
    }
    \subfloat[CPR=0.6, CIFAR-10]{
        \includegraphics[width=.33\linewidth]{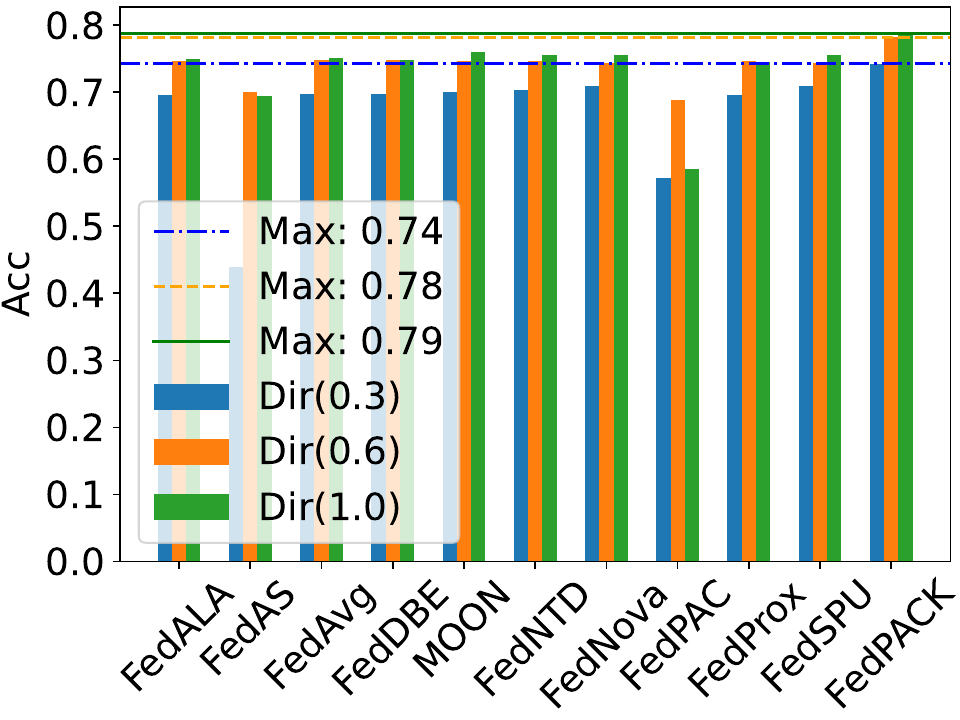}
        \label{app-fig4:c8}
    }
    \subfloat[CPR=1.0, CIFAR-10]{
        \includegraphics[width=.33\linewidth]{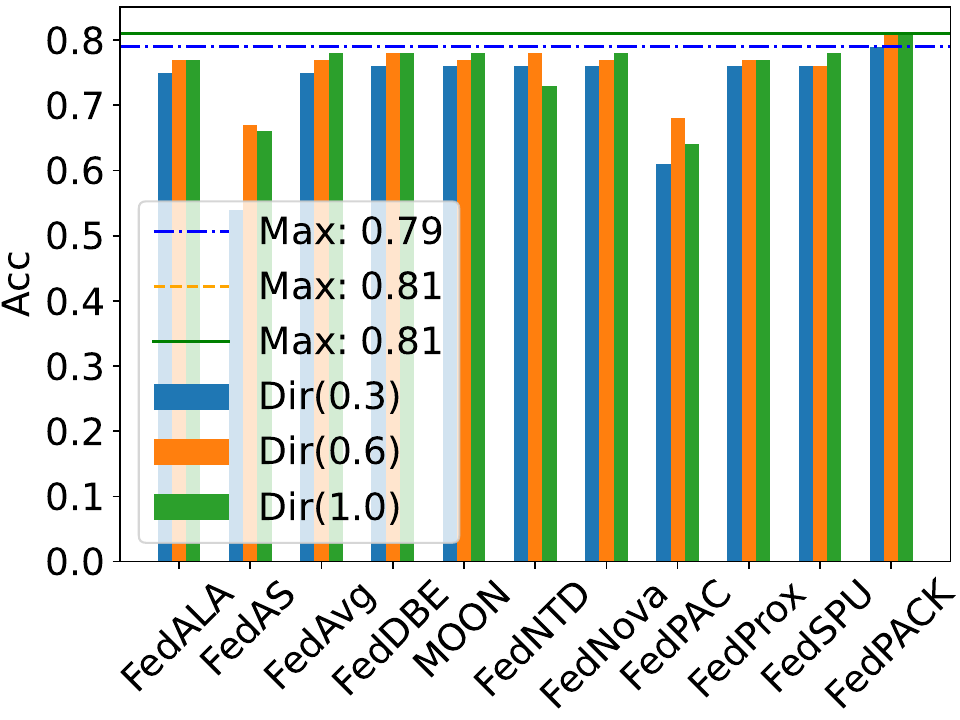}
        \label{app-fig4:c9}
    }
  \caption{The influence of the limited resources client participation ratio (CPR) on FMNIST, EMNIST, and CIFAR-10, where CPR represents the ratio of the number of clients per round to the total number of clients. The horizontal line indicates the accuracy of the FedCSPACK.Use the CNN model for testing}
    \label{app-fig4}
\end{figure*}

\section*{Appendix D}
\textbf{Numbers of Limited Resources Clients}.

\noindent Considering that clients can be disconnected due to resource constraints, figs.~\ref{app-fig42} and ~\ref{app-fig4} illustrate how client engagement success impacts model performance and stability. In figs.~\ref{app-fig42}, at CPR = 0.6, most methods (such as FedAvg, FedPAC, and FedProx) experienced a significant drop in accuracy on CIFAR-10 and CIFAR-100, even falling below 0.1. In contrast, FedCSPACK demonstrated higher accuracy across different datasets: 0.71 on FMNIST, 0.50 on EMNIST, 0.49 on CIFAR-10, and 0.14 on CIFAR-100. When the CPR = 1.0, overall accuracy generally improved, with FedCSPACK showing further improvement, reaching 0.76 on FMNIST, 0.65 on EMNIST, 0.54 on CIFAR-10, and 0.17 on CIFAR-100. These results indicate that data heterogeneity has a more significant impact on the model aggregation process when client engagement is low. Traditional methods, relying on all or randomly sampled client updates, are prone to the ``minority-dominated'' problem. Conversely, FedCSPACK, by employing a dual-weight masking mechanism and packet-level sparse communication, effectively filters high-quality parameter updates and suppresses noise interference, maintaining performance stability even at low participation rates. This demonstrates FedCSPACK's superior ability to handle client disconnections and data heterogeneity challenges.

In fig.~\ref{app-fig4}, FedCSPACK demonstrates superior performance on three typical datasets (FMNIST, EMNIST, and CIFAR-10), with its histogram length exhibiting significantly less fluctuation than some other models, indicating good stability even with limited resource participation. Even under extreme conditions of extremely high heterogeneity ($Dir(0.3)$) and extremely low client participation (CPR=0.3), FedCSPACK maintains the best accuracy, significantly outperforming state-of-the-art methods. Its superior performance stems from the proposed dual-weight masking mechanism and packet-level sparse communication strategy, which effectively achieves high-quality updates and maintains model consistency and stability under the dual challenges of limited resources and highly heterogeneous data. This demonstrates that FedCSPACK is not only suitable for standard federated learning scenarios but also has practical deployment value.

\begin{figure*}[htbp]
    \centering
    \subfloat[CIFAR-100,ACC]{
        \includegraphics[width=.24\textwidth]{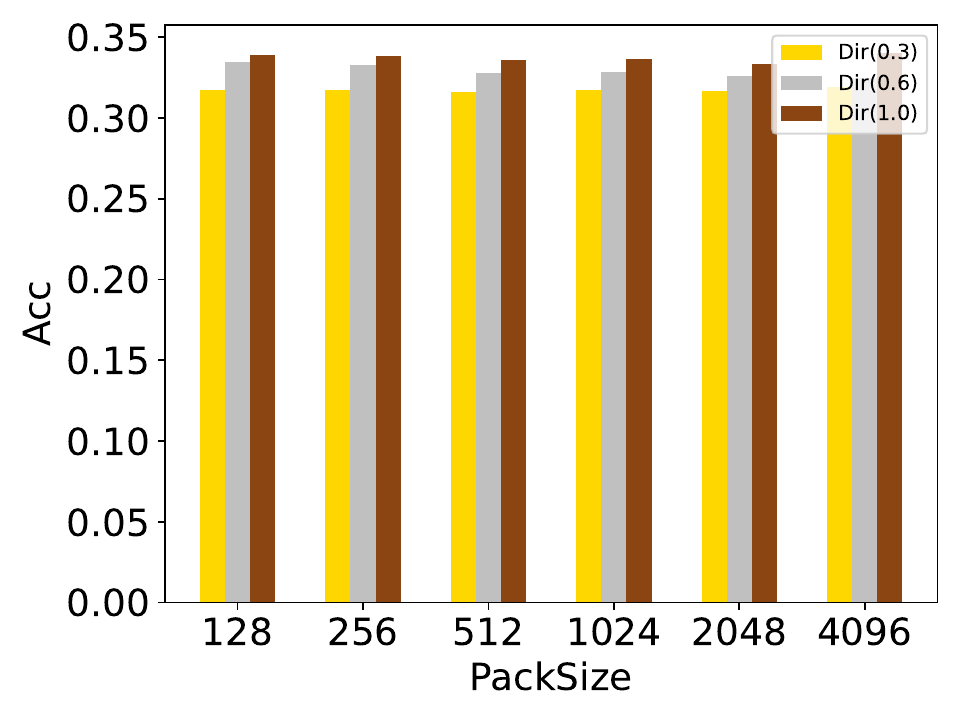}
        \label{app-fig3:cl1}
    } 
    \subfloat[CIFAR-100,TIMES]{
        \includegraphics[width=.24\textwidth]{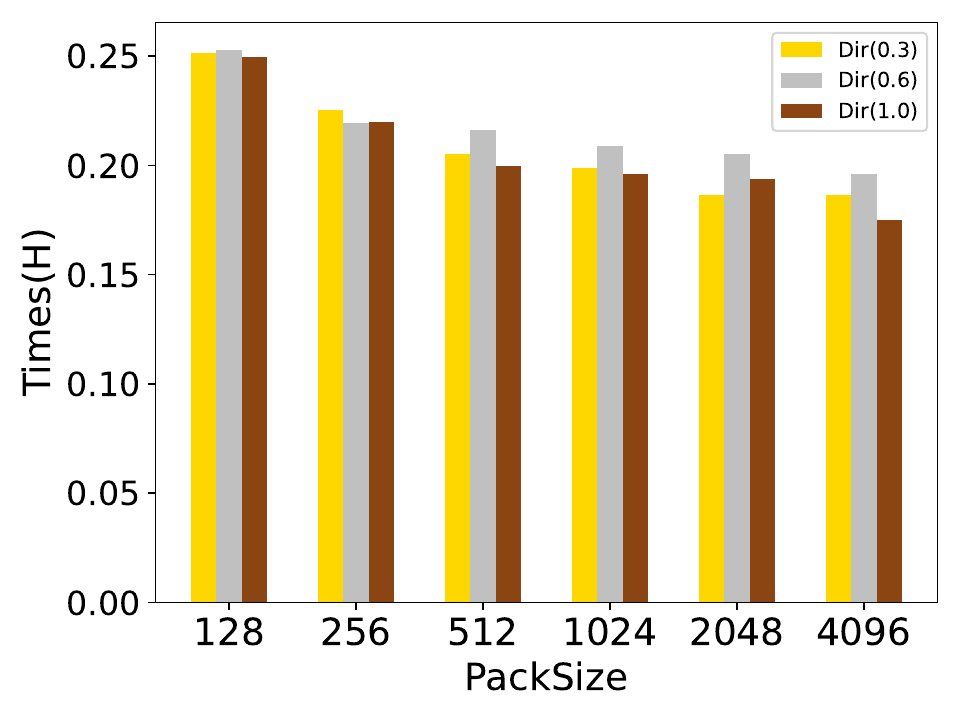}
        \label{app-fig3:cl2}
    }
    \subfloat[FMNIST,ACC]{
        \includegraphics[width=.24\textwidth]{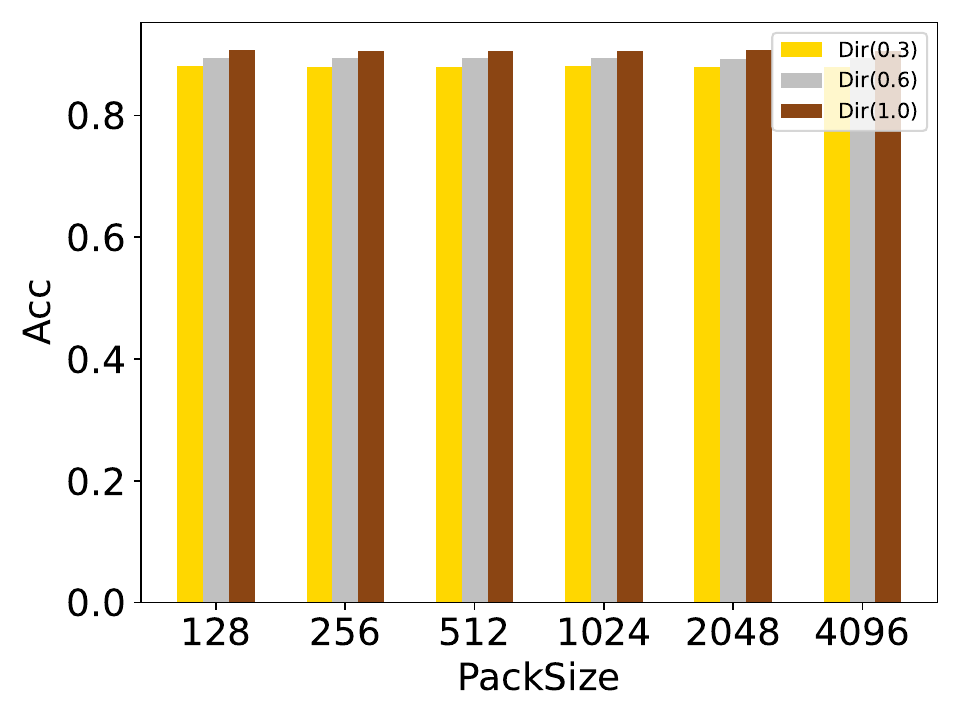}
        \label{app-fig3:cl3}
    }
    \subfloat[FMNIST,TIMES]{
        \includegraphics[width=.24\textwidth]{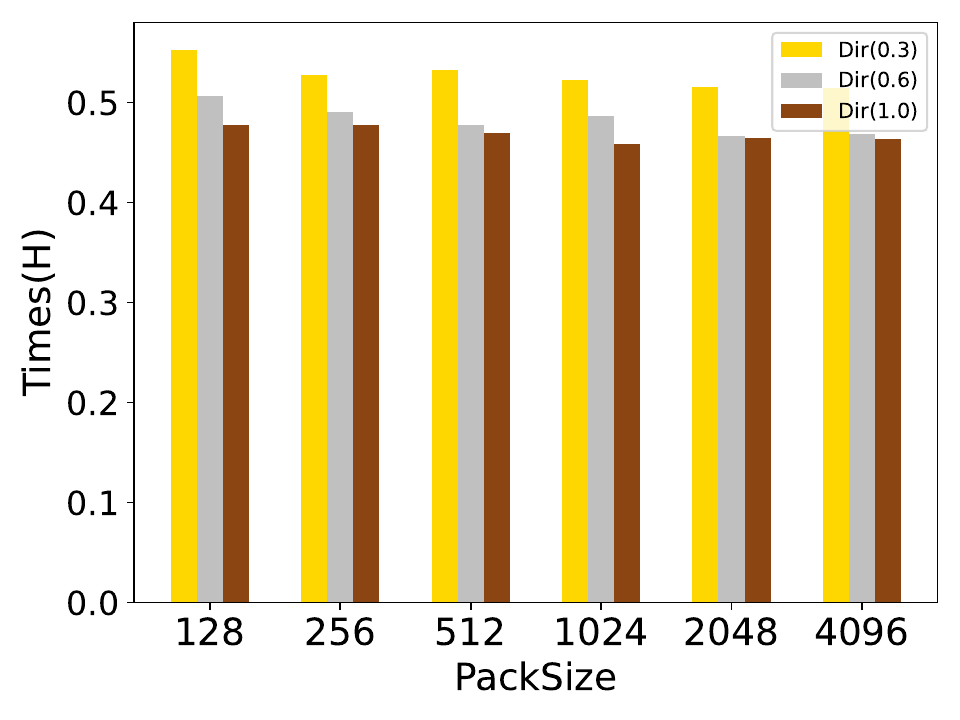}
        \label{app-fig3:cl4}
    }
    \caption{The impact of the $PACK$ size on model performance and computation time. Use the CNN model for testing}
    \label{app-fig3}
\end{figure*}
\section*{Appendix E}

\textbf{The Effect of the $PACK$ Size}. 

\noindent We further verify the effect of PACK on model performance and computation time in Fig. \ref{app-fig3}. As shown in Fig. \ref{app-fig3:cl1} and Fig. \ref{app-fig3:cl3}, on the three Dirichlet data heterogeneities, although the model accuracy improves with the increase of $PACK$ size, the extent is limited. Especially when the $PACK$ size exceeds 512, the improvement in accuracy almost comes to a standstill. This indicates that whether it is a complex or simple dataset, its sensitivity to the size of the $PACK$ is not high. In Fig. \ref{app-fig3:cl2} and Fig. \ref{app-fig3:cl4}, the computing time of CIFAR-100 decreases with the increase of $PACK$ size, while the reduction in computing time of FMNIST is relatively small. This indicates that under different $PACK$ sizes, FedCSPACK can effectively enhance the processing efficiency of complex datasets.

To sum up, the impact of $PACK$ size on model accuracy is limited, especially when the $PACK$ size exceeds 512; the improvement in accuracy hardly increases any further. However, the size of the $PACK$ has a significant impact on the computing time. Especially when the $PACK$ size exceeds 512, the computing time will decrease significantly. Therefore, in practical applications, the most appropriate $PACK$ size should be selected based on available computing resources and the complexity of the dataset to achieve the optimal balance between model performance and computing resource usage.

\end{document}